\documentclass{article}

\usepackage{iclr2027_conference,times}

\usepackage{amsmath,amsfonts,bm}

\def\eqref#1{equation~\ref{#1}}

\def\1{\bm{1}}

\DeclareMathAlphabet{\mathsfit}{\encodingdefault}{\sfdefault}{m}{sl}
\SetMathAlphabet{\mathsfit}{bold}{\encodingdefault}{\sfdefault}{bx}{n}

\newcommand{\R}{\mathbb{R}}

\usepackage[T1]{fontenc}
\usepackage[utf8]{inputenc}
\usepackage{graphicx}
\usepackage{wrapfig}
\usepackage{needspace}
\usepackage{booktabs}
\usepackage[table]{xcolor}
\usepackage{arydshln}
\usepackage{listings}
\usepackage{microtype}
\usepackage[most]{tcolorbox}
\usepackage{inconsolata}
\usepackage{enumitem}
\usepackage{pifont}
\usepackage{amssymb}
\usepackage{caption}
\usepackage{hyperref}
\usepackage{url}

\definecolor{expertgreen}{RGB}{0, 150, 0}
\definecolor{expertred}{RGB}{200, 0, 0}
\definecolor{lvlElem}{HTML}{9ECAE1}
\definecolor{lvlOlympiad}{HTML}{E6550D}
\definecolor{lvlUndergrad}{HTML}{74C476}
\definecolor{lvlResearch}{HTML}{6A51A3}
\newcommand{\lvlbox}[2]{%
  \begingroup
  \setlength{\fboxsep}{0pt}%
  \colorbox{#1!25!white}{\textcolor{black}{\makebox[1.15em][c]{\rule{0pt}{0.75em}\textbf{#2}}}}%
  \endgroup
}

\definecolor{citecolor}{HTML}{0071bc}
\definecolor{cadmiumgreen}{rgb}{0.0, 0.42, 0.24}
\definecolor{cornellred}{rgb}{0.7, 0.11, 0.11}
\hypersetup{
  colorlinks = true,
  linkcolor  = cornellred,
  citecolor  = citecolor,
  urlcolor   = cadmiumgreen,
}

\newtcolorbox{promptbox}{
  colback=gray!5, colframe=black!40,
  boxrule=0.4pt, arc=2pt,
  left=6pt, right=6pt, top=5pt, bottom=5pt,
  breakable,
  fontupper=\ttfamily\small
}

\newcommand{\dataset}{\textsc{ResearchMath-14k}}
\newcommand{\datareasoning}{\textsc{ResearchMath-Reasoning}}

\title{
\dataset{}: Scaling Research-Level Mathematics via Agents
}

\author{%
Guijin Son$^{1,2}$\hspace{0.5em}Seungyeop Yi$^{1}$\hspace{0.5em}Minju Gwak$^{3}$\hspace{0.5em}Hyunwoo Ko$^{2}$\hspace{0.5em}Wongi Jang$^{1}$\hspace{0.5em}Youngjae Yu$^{1}$\thanks{Corresponding author.}\\[4pt]
\normalfont $^{1}$Seoul National University \quad $^{2}$OneLineAI \quad $^{3}$Yonsei University\\[4pt]
\normalfont \texttt{guijin.son@snu.ac.kr} \quad \texttt{youngjaeyu@snu.ac.kr}
}
\hypersetup{
  pdftitle={ResearchMath-14k: Scaling Research-Level Mathematics via Agents},
  pdfauthor={Guijin Son, Seungyeop Yi, Minju Gwak, Hyunwoo Ko, Wongi Jang, Youngjae Yu}
}

\iclrfinalcopy 

\begin{document}
\maketitle
\pagestyle{fancy}
\lhead{Preprint}

\begin{abstract}

The frontier of mathematics is defined by problems whose solutions are not yet known. However, whether language models can meaningfully engage with such problems without human intervention remains unclear. A major obstacle is the lack of large-scale research-level math datasets. To this end, we introduce \dataset{}, a set of $14{,}056$ problems curated from academic sources via a multi-agent pipeline. \dataset{} spans 11 mathematical domains and ranks above existing math datasets on knowledge, novelty, and procedural difficulty. To our knowledge, it is the largest research-level mathematical problem set available for training. We additionally generate $220$K teacher trajectories through targeted prompting, followed by behavioral filtering. Notably, however, generating correct trajectories is nontrivial at this level, and two LLM judges label only $3.7\%$ and $4.3\%$ of sampled \textsc{ResearchMath} training trajectories as correct. Nevertheless, across three model families, full-parameter training on \textsc{ResearchMath} improves performance on graduate- and research-level mathematics benchmarks by $2.1$ points over the starting checkpoints. In comparison, training on existing datasets such as DASD and \texttt{Nemotron-SFT-Math-v4} changes performance by $0.0$ and $-0.5$ points, respectively. Notably, mixing DASD with \textsc{ResearchMath} yields higher scores than token-matched DASD alone on benchmarks covering olympiad short-form ($+2.0$), graduate- and research-level short-form ($+0.8$), graduate- and research-level symbolic ($+2.6$), and proof evaluation ($+5.7$). Further analysis suggests that research-level mathematical content and greater reasoning diversity may help explain why \textsc{ResearchMath} provides complementary supervision to contemporary datasets. We make \dataset{} publicly available for future works on research-level mathematical reasoning.\footnote{\url{https://huggingface.co/datasets/amphora/ResearchMath-14k}}

\end{abstract}

\section{Introduction}
\label{sec:intro}

Mathematicians are trained over years, progressing from undergraduate textbooks and exercises to seminar problems, qualifying exams, and short-term research, gradually acquiring the practices required to become researchers. Frontier proprietary models appear to internalize parts of this curriculum~\citep{alexeev2026short,alexeev2026short2,zheng2026ai}. However, the open-source landscape has not kept pace. Nearly all publicly available math training data targets contest-style problems at the olympiad level or below~\citep{numina_math_datasets,fan2025megascience}, and the few datasets that do reach the research frontier are positioned as held-out evaluation benchmarks, often gate-kept to prevent contamination~\citep{glazer2024frontiermath,phan2025humanity}. \emph{Where, then, can research-level mathematical questions be obtained at scale?} Recent work has largely relied on two expensive sources: multi-LLM pipelines that synthesize difficult problems~\citep{zhang2026realmath, dekoninck2026matharena}, or expert mathematicians who write by hand~\citep{son2026judging, garre2026riemann}. Both approaches are valuable, but neither provides an easy path to a broad, open training corpus. Instead, we leverage recent advances in agentic pipelines to recover research-level questions already embedded in the mathematical literature and transform them into self-contained training data.

In this work, we collect $1{,}233$ open-problem lists and research papers from arXiv, curated open-problem web pages, and academic repositories, using a multi-agent pipeline to identify candidate questions, reconstruct the necessary context, and normalize them into standalone research-level problems. The resulting dataset, \dataset{}, contains $14{,}056$ research-level mathematical questions across $11$ fields, including number theory, geometry and topology, combinatorics, and mathematical physics. In terms of \emph{difficulty}, it ranks roughly $400$ Elo points above \texttt{AceMath}~\citep{liu2025acemath}, \texttt{AIME}~\citep{aime24,aime25,aime26}, \texttt{HLE-Verified}~\citep{phan2025humanity,zhai2026hle}, and \texttt{NuminaMath}~\citep{numina_math_datasets} across knowledge, novelty, and procedural difficulty (Section~\ref{sec:dataset-statistics}). To assess \emph{quality}, four mathematics PhD students contributed $138$ retained reviews, of which $115$ ($83.3\%$) assigned the top rating on all five self-containedness dimensions (Section~\ref{sec:human-audit}). For distillation, we use \texttt{GPT-OSS-120B}~\citep{agarwal2025gpt} and \texttt{Qwen3-30B-A3B}~\citep{yang2025qwen3} to generate a collection of $220$K reasoning trajectories. However, it should be noted that ground-truth solutions are largely unavailable or nontrivial to collect at this scale. Therefore, rather than applying correctness filtering, we apply a behavior-based filter that removes non-attempts and fabricated citations. Interestingly, even without correctness filtering, \textsc{ResearchMath} yields the largest overall gain ($+4.0$ points) across three model families and five benchmarks, outperforming \texttt{DASD}~\citep{yan2026distribution} and \texttt{Nemotron-SFT-Math-v4}~\citep{nvidia2026nemotronsftmathv4}. The advantage is particularly pronounced on graduate- and research-level benchmarks, with a $2.1$-point gain versus $0.0$ and $-0.5$ for the two comparison datasets. Moreover, combining \textsc{ResearchMath} with DASD improves all three models across four benchmark categories compared with token-matched DASD alone. These findings suggest that unverified research-level attempts provide useful supervision beyond existing training data. Further analysis finds greater embedding diversity in \textsc{ResearchMath} trajectories than in the baseline datasets, suggesting that varied reasoning content may contribute to these benefits.

\begingroup
\setlength{\parskip}{2pt}
\noindent Our contributions are threefold:
\setlength{\leftmargini}{1.3em}
\begin{itemize}[topsep=1pt,partopsep=0pt]
    \item We present a multi-agent pipeline that recovers research-level mathematical questions from fragmented literature and converts them into self-contained training problems.
    \item We introduce \dataset{}, to our knowledge the largest collection of research-level mathematical problems, along with $220$K reasoning trajectories.
    \item We show that behavior-filtered, unverified research-level trajectories improve reasoning across three model families and complement existing post-training data. We further investigate mathematical content and reasoning diversity as possible explanations for these gains.
\end{itemize}
\endgroup

\Needspace{7\baselineskip}

\section{Related Works}

\begin{wraptable}{R}{0.58\textwidth}
\vspace{-13pt}
\centering
\scriptsize
\setlength{\tabcolsep}{3pt}
\caption{\footnotesize \textbf{Representative public math datasets.} Top section: training datasets are large, but do not expand to the research level. Lower section: existing research-grade resources are small ($<\,3$k items) and evaluation-only. \dataset{} fills both gaps, large-scale and research-grade. Difficulty key: \lvlbox{lvlElem}{G}~grade-school, \lvlbox{lvlOlympiad}{O}~olympiad, \lvlbox{lvlUndergrad}{U}~undergrad, \lvlbox{lvlResearch}{R}~research.}
\label{tab:dataset_compare}
\begin{tabular}{lrlc}
\toprule
Dataset & \#Problems & Source & Diff. \\
\midrule
GSM8K~\citep{cobbe2021training}                              & $8.5$k        & textbook        & \lvlbox{lvlElem}{G}      \\
MATH~\citep{hendrycks2021measuring}                            & $12.5$k       & competitions    & \lvlbox{lvlOlympiad}{O}  \\
LeanDojo~\citep{yang2023leandojo}                         & $122$k        & research lit.\  & \lvlbox{lvlUndergrad}{U} \\
MathInstruct~\citep{yue2024mammoth}                       & $262$k        & synthetic       & \lvlbox{lvlUndergrad}{U} \\
MetaMathQA~\citep{yu2024metamath}                         & $395$k        & synthetic       & \lvlbox{lvlOlympiad}{O}  \\
PRM800K~\citep{lightman2024let}                        & $800$k        & competitions    & \lvlbox{lvlOlympiad}{O}  \\
NuminaMath~\citep{numina_math_datasets}                   & $860$k        & synthetic       & \lvlbox{lvlOlympiad}{O}  \\
AceMath-Instruct~\citep{liu2025acemath}                   & $1.66$M       & synthetic       & \lvlbox{lvlOlympiad}{O}  \\
OpenMathInstruct~\citep{toshniwal2024openmathinstruct}    & $1.8$M        & synthetic       & \lvlbox{lvlOlympiad}{O}  \\
\midrule
Riemann-Bench~\citep{garre2026riemann}                    & $25$          & expert  & \lvlbox{lvlResearch}{R}  \\
FrontierMath~\citep{glazer2024frontiermath}               & $300$         & expert  & \lvlbox{lvlResearch}{R}  \\
Soohak~\citep{son2026soohak}                              & $439$         & expert  & \lvlbox{lvlResearch}{R}  \\
GHOSTS~\citep{frieder2023mathematical}                          & $709$         & textbook        & \lvlbox{lvlResearch}{R}  \\
HARDMath~\citep{fan2025hardmath}                          & $1{,}426$     & textbook        & \lvlbox{lvlResearch}{R}  \\
HLE~\citep{phan2025humanity}                                   & $2{,}500$     & expert  & \lvlbox{lvlResearch}{R}  \\
\midrule
\textbf{\dataset{}} & $\mathbf{14{,}056}$ & \textbf{research lit.} & \lvlbox{lvlResearch}{R} \\
\bottomrule
\end{tabular}
\vspace{-8pt}
\end{wraptable}

\subsection{Mathematical Post-Training Data for LLMs}
\label{sec:related-math-posttraining}

Mathematical post-training is central to drawing forth general reasoning capabilities in LLMs. Accordingly, substantial effort has gone into building post-training data. Earlier work has evolved from collecting solved grade-school and competition problems with answer or process annotations~\citep{cobbe2021training,hendrycks2021measuring,lightman2024let}, through aggregating and standardizing existing problem-solution resources~\citep{yue2024mammoth,numina_math_datasets} and extracting formal theorem libraries~\citep{yang2023leandojo}, to using LLMs to rewrite questions and synthesize solutions~\citep{yu2024metamath,liu2025acemath,toshniwal2024openmathinstruct}. Unfortunately, these strategies do not readily extend to research-level mathematics, where large collections of problems paired with solutions are scarce. Suitable questions arise only sporadically within specialized work and often require expert annotation~\citep{abouzaid2026first}. Existing research-level resources therefore remain small and evaluation-oriented, including \texttt{FrontierMath}~\citep{glazer2024frontiermath}, \texttt{Riemann-Bench}~\citep{garre2026riemann}, and \texttt{SOOHAK}~\citep{son2026soohak}. One exception is \texttt{RealMath}~\citep{zhang2026realmath}, which uses automated collection to assemble research-level mathematical problems. Meanwhile, recent reports describe frontier AI systems resolving long-standing open problems across several mathematical fields, including Erd{\H{o}}s questions~\citep{zheng2026ai,alexeev2026short,alexeev2026short2}, the existence of non-sofic groups, and Ehrhart's volume conjecture~\citep{openai2026tenadvances}. However, no public resource combines research-level difficulty with the scale needed for open models to keep pace (Table~\ref{tab:dataset_compare}).

\newpage

\subsection{Agentic Data Collection}
\label{sec:related-agentic-data}
Although expert curation offers direct control over training data, maintaining consistent quality requires costly review and makes the process difficult to scale. These constraints have motivated longstanding work on scalable data construction. Earlier methods relied mainly on prompt-based generation, either rewriting and expanding seed examples~\citep{yu2024metamath,toshniwal2025openmathinstruct} or grounding new questions in source documents~\citep{yuan2026naturalreasoning}. More recent work has introduced agentic pipelines that search external sources and iteratively propose and refine tasks~\citep{liu2025webexplorer}. Unlike one-shot LLM generation, which typically assumes a prepared input context, agentic pipelines can navigate long documents, move across related sources, and search the web for information missing from the initial input. These capabilities are particularly useful for research mathematics, where there is no pool of cleanly stated problems comparable to textbook and olympiad collections, and candidate questions must instead be recovered from papers and their surrounding literature. Accordingly, in this work, we introduce a domain-specific adaptation of agentic data collection for research mathematics. Our pipeline recovers researcher-authored open problems, reconstructs omitted context, checks their resolution status, and converts them into self-contained training examples, addressing the scarcity and contextual fragmentation that hinder scalable data construction at the research level.

\begin{figure}[t]
  \centering
  \includegraphics[width=0.96\textwidth]{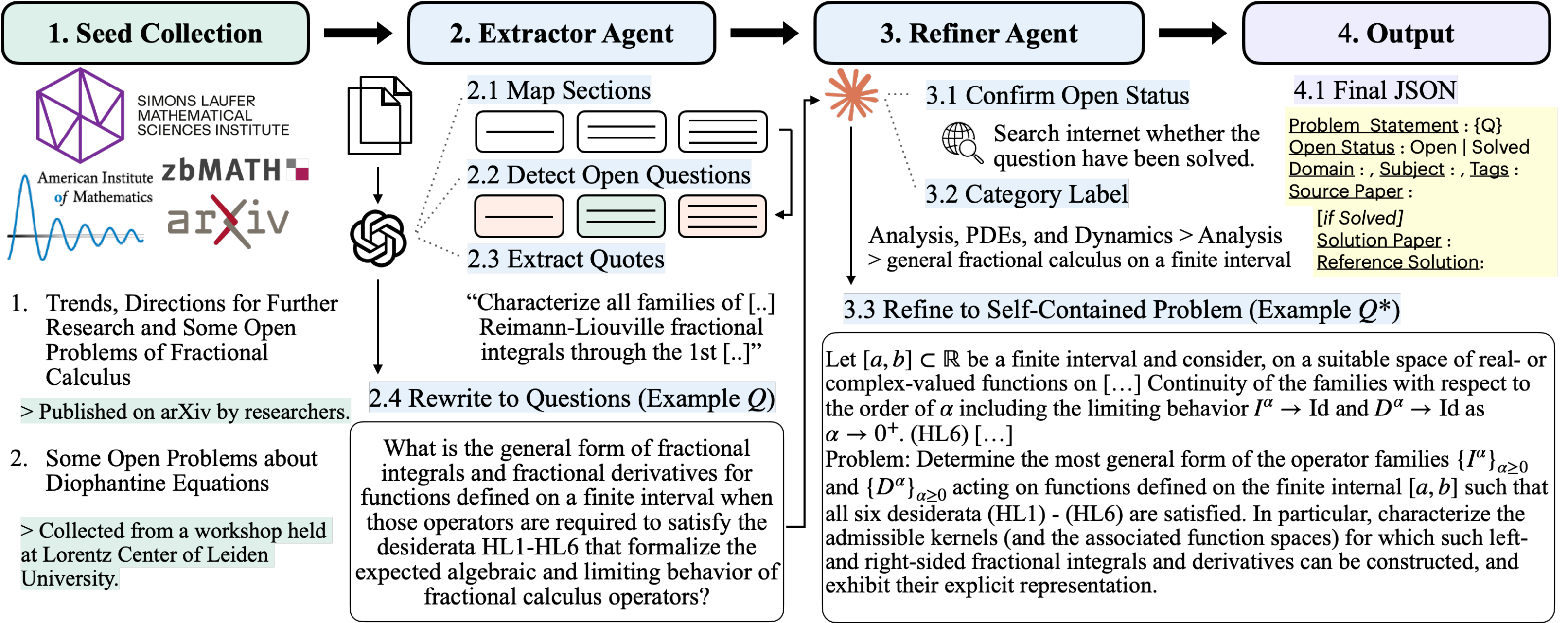}
  \caption{\footnotesize \textbf{Agentic construction pipeline for \dataset{}.} Starting from curated open-problem lists and research papers, an extractor agent maps document sections, detects candidate open questions, preserves supporting quotes, and rewrites them as standalone questions. A refiner agent then verifies open status, assigns taxonomy labels, and rewrites each candidate into a self-contained research problem before producing the final JSON record with statement, status, domain metadata, source, and solution fields when applicable.}
  \label{fig:collection_pipeline}
\end{figure}

\section[ResearchMath-14k: Collecting Research-Level Math Problems]{\dataset{}: Collecting Research-Level Math Problems}
\label{sec:collection}

In this section, we describe how we collect and refine existing open questions into self-contained problems (Section~\ref{sec:collection-process}) and remove near duplicates and potential evaluation contamination (Section~\ref{sec:filtering}). We then introduce the resulting dataset through analyses of its domain, source, status, and difficulty (Section~\ref{sec:dataset-statistics}). Finally, Section~\ref{sec:human-audit} presents findings from a human expert review.

\subsection{Collecting Existing Open Questions}
\label{sec:collection-process}
\dataset{} is constructed using a two-stage pipeline: an Extractor agent pulls candidate statements from each source, and a Refiner agent rewrites them into a self-contained problem, consulting online references. This produces $20{,}835$ problems from $1{,}233$ documents, see Figure~\ref{fig:collection_pipeline}.

\paragraph{Sources.}~Mathematicians have long published unresolved questions through workshops, surveys, and curated lists, both to attract collaborators and to record which questions a field regards as important enough to foreground for the broader community. Our pipeline captures both classical entries, such as Hilbert- or Erd{\H{o}}s-style problem lists, and \emph{contemporary} problem statements about modern mathematical objects and local technical settings. The latter are closer to the day-to-day research questions a working mathematician might pose at a workshop or in a recent survey, and are therefore the kind of supervision signal we target. See Appendix~\ref{app:source-example-comparison} for examples. Specifically, source documents are drawn from three streams. \textbf{arXiv open-problem papers} (e.g., ~\citet{diethelm2022trends}) ($524$ documents, $8{,}182$ problems) are surveyed by searching arXiv for titles and abstracts mentioning ``open problems,'' or ``unsolved.''  \textbf{Open-problem web pages} ($161$ documents, $5{,}331$ problems) are discovered by Google search and cover hosts such as \texttt{academia.edu}, MathOverflow, and Wikipedia. \textbf{Problem-session sheets and curated lists} ($548$ documents, $7{,}322$ problems) are the third stream and include two sub-types: AIM-style \emph{workshop problem sessions} where participants pose questions at the end of a meeting,\footnote{\url{https://aimath.org/pastworkshops/nonselfadjointproblems.pdf}} and \emph{conference/proceedings open-problem rounds} compiled by an editor.

\paragraph{Extractor Agent.}

The Extractor, driven by Codex with GPT-5.5 at \emph{xhigh} reasoning effort, processes one source per run. It first follows the source URL down to the PDF or HTML page that holds the full text, \textbf{discarding any document hidden behind a paywall or a non-permissive license}. Before extraction, it skips papers that do not pose open problems (e.g., papers that merely mention ``open problem''). It then reads the paper end-to-end and extracts each open problem as a verbatim quote with a first-level rewrite. While rewriting, the model is instructed to jump back and forth through the paper to pull in every definition and statement needed to understand that problem. Across the $1{,}233$ documents the Extractor yields a mean of $16.9$ questions per source (median $10$, maximum $358$).

\paragraph{Refiner Agent.}

Reading through the extracted questions, the authors noticed that some of the snippets still miss the definitions, notation, and hypotheses the original paper treats as already given. The Refiner, driven by Claude Code with Opus 4.7 at \emph{medium} reasoning effort, fills that gap.  It performs two tasks. First, it re-reads the original paper to inline every definition and hypothesis needed to state the problem in isolation. Second, it searches up to ten later papers that cite or extend the source, both to pull in the background of the source treated as implicit and to determine whether the problem has since been resolved. We audit $500$ random records labels using GPT-5.5 as LLM-Judge. It labels $94.2\%$ of refined statements as self-contained, compared with $67.2\%$ of original extractions, a $27.0$ percentage-point improvement (Appendix~\ref{app:self-containment-audit}). Refined statements also average $1{,}192$ characters, up from $290$ at the Extractor stage, a $4.1\times$ expansion.

\subsection{Filtering Near Duplicates}
\label{sec:filtering}
The collection pipeline produces a seed set of $20{,}835$ problems, but multiple sources often state the same open problem in slightly different forms, making duplicate filtering necessary. We embed all problems with \texttt{Qwen3-Embedding-8B}~\citep{zhang2025qwen3} and compute pairwise similarities over both the original statements and the self-contained rewrites. Questions extracted from the same paper often share extensive background text and can look similar even when they are distinct, so a low similarity threshold would introduce many false positives. After manually inspecting borderline pairs at several cutoffs, we set the threshold to $0.9$. A pair is marked as a duplicate if either similarity score exceeds this value. For duplicates, we deprioritize sources found through Google search and choose at random when sources have equal priority. This leaves a final collection of $14{,}056$ problems. See Appendix~\ref{app:filtering-details} for the similarity distribution and non-duplicate pairs near the threshold. For decontamination, we check every problem for $14$- and $18$-gram overlap with \texttt{AIME} (2024--2026), \texttt{HLE}, \texttt{SOOHAK}, \texttt{ArxivMath}~\citep{dekoninck2026beyond}, and \texttt{QEDBench}~\citep{gonzalez2026qedbench} (Appendix~\ref{app:decontamination}).

    \subsection{Dataset Statistics}
    \label{sec:dataset-statistics}

\paragraph{Composition.}
Each problem is organized within a three-level taxonomy that progresses from a broad domain to a
macro-subject and finally to a research-level category tag. The taxonomy comprises $11$ domains,
$28$ macro-subjects, and $11{,}611$ unique tags. The display below presents the domain distribution
and one representative branch. The corpus covers a broad range of fields, with four large areas accounting for $8{,}971$ problems ($63.82\%$): Analysis/PDEs/Dynamics, Mathematical Physics, Discrete Mathematics/Combinatorics, and Geometry/Topology. 

\noindent
\begin{minipage}[t]{0.54\textwidth}
\vspace{0pt}
\centering
\resizebox{0.93\linewidth}{!}{%
$\displaystyle
\setlength{\arraycolsep}{2pt}
\mathcal{G}=\left\{
\begin{array}{l}
\text{Analysis, PDEs, and Dynamics (22.7\%);}\\
\text{Mathematical Physics (14.4\%);}\\
\text{Discrete Mathematics and Combinatorics (13.5\%);}\\
\text{Number Theory (5.7\%);}\\
\text{Geometry and Topology (13.1\%);}\\
\text{Theoretical Computer Science (5.3\%);}\\
\text{Algebra and Representation Theory (9.2\%);}\\
\text{Probability, Statistics, and ML (4.5\%);}\\
\text{Applied and Computational Mathematics (6.0\%);}\\
\text{Logic and Foundations (3.2\%);}\\
\text{Other / Cross-disciplinary (2.2\%).}
\end{array}
\right.
$%
}
\end{minipage}\hfill
\begin{minipage}[t]{0.43\textwidth}
\vspace{0pt}
\centering
{\fontsize{8.3}{9.3}\selectfont
$\displaystyle
\setlength{\arraycolsep}{1pt}
\begin{array}{rcl}
\mathrm{domain} & = & \text{Geometry and Topology}\\
\mathrm{macro} & = & \text{Algebraic Geometry}\\
\mathrm{tags} & = &
\left\{
\begin{array}{l}
\text{Hilbert schemes;}\\
\text{Brill-Noether theory;}\\
\text{curves over finite fields;}\\
\text{Kuznetsov components;}\\
\text{stability conditions;}\\
\text{line arrangements;}\\
\text{affine spaces;}\\
\text{Hurwitz spaces;}\\
\text{automorphism groups of curves.}
\end{array}
\right.
\end{array}
$%
}
\end{minipage}
\par

\noindent A small fraction, $311$ problems ($2.2\%$), falls into the \emph{Other/Cross-disciplinary} group and covers science-adjacent open questions (e.g., on supernova progenitors, origin of language, computational theory of mind). Open problems form the majority ($8{,}313$, $59.14\%$), followed by \emph{unknown} ($2{,}489$, $17.71\%$), \emph{partially solved} ($2{,}083$, $14.82\%$), and \emph{solved} ($1{,}171$, $8.33\%$). The set is source-diverse, spanning $1{,}138$ unique documents, with the top $10$ contributing $1{,}431$ problems ($10.18\%$).

\begin{wrapfigure}{r}{0.5\textwidth}
\centering
\vspace{-5pt}
\includegraphics[width=\linewidth]{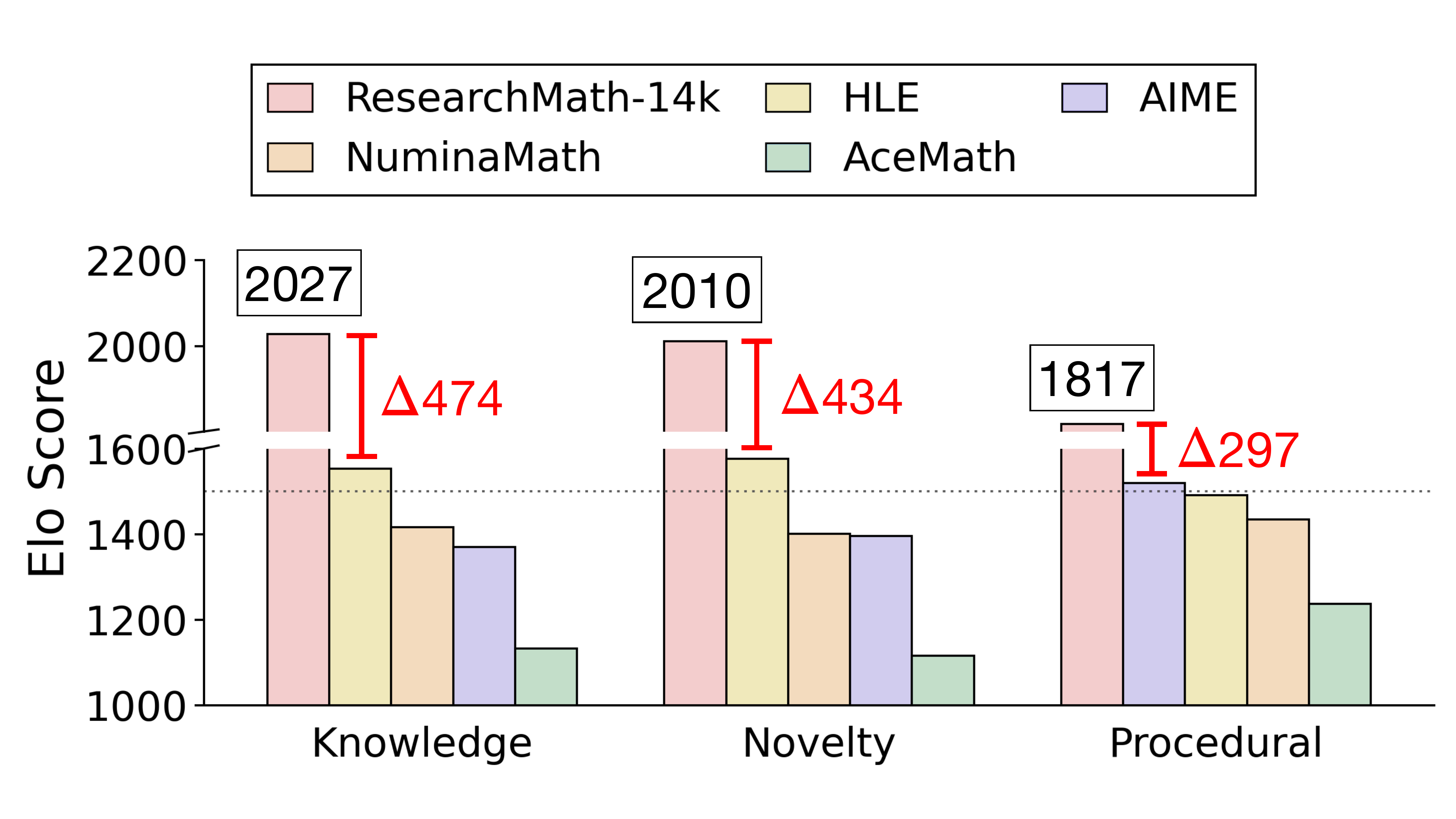}
\caption{\footnotesize \textbf{Elo Ratings for Difficulty Comparison.} Ratings are computed from pairwise LLM difficulty judgments. All sources start at $1500$ with $k=32$; wins score $1$, losses score $0$, and draws score $0.5$. Higher Elo means the source is judged more difficult more often.}
\label{fig:elo_score}
\end{wrapfigure}

\paragraph{Difficulty.}~Difficulty is multidimensional, and a problem can be hard because it requires familiarity with specialized mathematical concepts (\textit{Knowledge}), demands novel thinking that deviates from existing approaches (\textit{Novelty}), or involves compute-heavy multi-step reasoning (\textit{Procedural}). We compare \dataset{} against \texttt{AceMath}~\citep{liu2025acemath}, \texttt{AIME(2024--2026)}, \texttt{HLE-Verified}~\citep{zhai2026hle}, and \texttt{NuminaMath}~\citep{numina_math_datasets}. From each of the five datasets we sample $90$ problems and consider all $\binom{5}{2}$ dataset pairs. For each pair we randomly draw $100$ cross-dataset problem pairs and randomize their order, giving $1{,}000$ total comparisons. Each comparison is judged by \texttt{GPT-5-mini} along the three axes, producing win/loss/draw labels from which we compute Elo ratings. On all three axes, \dataset{} ranks above these existing math datasets by roughly $400$ Elo points (Figure~\ref{fig:elo_score}), suggesting that \textsc{ResearchMath} presents qualitatively more demanding mathematical problems. Additionally, motivated by \citet{ballon2026relationship}, we compare completion lengths across the five datasets using two LLMs, with problem samples matched by statement length. Both models produce completions on \dataset{} comparable in length to those on \texttt{HLE-Verified} and substantially longer than the rest. This provides complementary evidence that \dataset{} poses greater difficulty for the evaluated models. See Appendix~\ref{app:length-controlled-difficulty} for further details and results.

\subsection[Human Expert Review]{Human Expert Review}
\label{sec:human-audit}

To further validate the quality of the questions, we hire four mathematics PhD students whose expertise spans number theory and arithmetic geometry, algebraic geometry, topology, and dynamics. Each expert was provided with a candidate pool of roughly $1{,}000$ problems and instructed to choose those they felt familiar enough to review. For each selected problem, the expert wrote an opinion, completed an error-note field by recording a concern or stating that none was found, optionally suggested a fix, and rated notation, terminology, assumptions and well-posedness, goal clarity, and freedom from dangling references, assuming the intended reader was a specialist in the relevant field. Experts were compensated at \$26 per review. After retaining the latest submission for each reviewer and problem identifier, the study contains $138$ reviews. These correspond to a total of \$3,588.

\begin{table}[t]
\centering
\begin{minipage}[t]{0.48\textwidth}
\centering
\footnotesize
\setlength{\tabcolsep}{1.8pt}
\caption{\footnotesize \textbf{Results of the human expert review of \dataset{}.} Each review rates five dimensions of statement self-containedness from $1$ to $5$. The reviewers used only scores $1$, $3$, and $5$.}
\label{tab:human-expert-audit}
\begin{tabular*}{\linewidth}{@{\extracolsep{\fill}}lcc@{}}
\toprule
Dimension & Mean & Score $5$ \\
\midrule
Q1~~Notation & $4.91$ & $95.7\%$ \\
Q2~~Terminology & $4.88$ & $94.2\%$ \\
Q3~~Assumptions and well-posedness & $4.84$ & $93.5\%$ \\
Q4~~Goal clarity & $4.80$ & $92.8\%$ \\
Q5~~No dangling references & $4.96$ & $98.6\%$ \\
\bottomrule
\end{tabular*}

\end{minipage}\hfill
\begin{minipage}[t]{0.495\textwidth}
\centering
\footnotesize
\setlength{\tabcolsep}{1pt}
\setlength{\dashlinedash}{0.4pt}
\setlength{\dashlinegap}{1.5pt}
\renewcommand{\arraystretch}{0.95}
\caption{\footnotesize \textbf{LLM-judge verdicts (\%).} Each judge assesses $1{,}000$ trajectories per dataset. Paired columns show Fable-5.1 (left) and GPT-5.6-Sol (right). Nemotron denotes \texttt{Nemotron-SFT-Math-v4}. See Appendix~\ref{app:approach-correctness-prompt} for LLM-Judge prompts.}
\label{tab:trajectory-judge-summary}
\fontsize{8}{10.5}\selectfont
\begin{tabular*}{\linewidth}{@{\extracolsep{\fill}}l>{\centering\arraybackslash}p{20pt}:>{\centering\arraybackslash}p{20pt}|>{\centering\arraybackslash}p{19pt}:>{\centering\arraybackslash}p{19pt}|>{\centering\arraybackslash}p{25pt}:>{\centering\arraybackslash}p{25pt}@{}}
\toprule
Verdict & \multicolumn{2}{c!{\vrule width\arrayrulewidth}}{Nemotron} & \multicolumn{2}{c!{\vrule width\arrayrulewidth}}{DASD} & \multicolumn{2}{c}{ResearchMath} \\
\midrule
Correct & $92.3$ & $77.8$ & $82.6$ & $72.5$ & $4.3$ & $3.7$ \\
Partial & $5.7$ & $16.4$ & $12.3$ & $14.3$ & $18.6$ & $7.1$ \\
Incorrect & $1.3$ & $5.7$ & $4.9$ & $13.2$ & $65.5$ & $84.2$ \\
Unverifiable & $0.7$ & $0.1$ & $0.3$ & $0.0$ & $11.6$ & $5.0$ \\
\bottomrule
\end{tabular*}

\end{minipage}
\end{table}

Table~\ref{tab:human-expert-audit} summarizes the expert ratings by dimension. Only $4$ of the $138$ retained reviews ($2.9\%$) assigned a score of $1$ on any dimension, while $115$ ($83.3\%$) assigned the top score on all five dimensions. Experts provided notes in $19$ reviews ($13.8\%$), but only $5$ ($3.6\%$) reported specific mathematical or statement-level defects that could mislead or block the intended reader. The remaining $14$ notes mainly offer suggestions for improving notation, presentation, and clarity. These results support the self-containedness of most reviewed statements and the suitability of \dataset{} as a source of research-level problems for training. Appendix~\ref{app:human-audit} details the response consolidation and limitations, with examples of the reported defects and positive opinions.

\section[Teacher Trajectory Generation and Filtering]{Teacher Trajectory Generation and Filtering}
\label{sec:traces}
\label{sec:generating-response}

For the problems collected in Section~\ref{sec:collection}, we use \texttt{GPT-OSS-120B} and \texttt{Qwen3-30B-A3B} to generate $220$K reasoning traces. In this section, we examine their correctness and behavioral limitations (Section~\ref{sec:results}) and describe the prompting and filtering used to prepare the training data (Section~\ref{sec:mitigation}).

\subsection[Correctness and Behavioral Limitations]{Correctness and Behavioral Limitations}
\label{sec:results}
\label{sec:experiment-setup}
\label{sec:baselines}

\paragraph{Mathematical correctness remains uncertain.}\label{sec:trajectory-correctness}
We assess $1{,}000$ trajectories each from \textsc{ResearchMath}, DASD~\citep{yan2026distribution}, and \texttt{Nemotron-SFT-Math-v4}~\citep{nvidia2026nemotronsftmathv4} using \texttt{GPT-5.6-Sol} and \texttt{Fable-5.1} (Table~\ref{tab:trajectory-judge-summary}). The two judges label only $3.7\%$ and $4.3\%$ of \textsc{ResearchMath} trajectories as correct, respectively, well below the rates for both baselines. They agree on $628$ incorrect and $25$ correct verdicts across $994$ jointly assessed \textsc{ResearchMath} trajectories. Although LLM judges may err, these estimates suggest that many sampled trajectories contain mathematical errors. These results are unsurprising, particularly given the difficulty of research-level problems for our open-weight teacher models. Appendix~\ref{app:correctness-diversity} details the judgments.
\paragraph{Teacher models often produce non-attempts or fabricated citations on open problems.}~Building on the LLM-based assessment, the authors manually review $100$ randomly sampled trajectories to examine teacher behavior in greater detail. During the review, we identified $25$ cases in which the teacher does not attempt to solve the problem. Of these, $21$ list related references and simply declare the problem open. The remaining four fake as if they have solved the problem by subtly changing the task, either by narrowing the original conditions or by presenting related results. The review also shows that models often use citations to make their responses appear grounded. In an early experiment described in Appendix~\ref{app:lora-results}, fine-tuning \texttt{Qwen3-4B} on the original, unfiltered teacher trajectories produced degenerate outputs and hallucinations. We suspect that the behaviors identified in our review contributed to this failure, motivating an additional post-processing step. Interestingly, these behaviors persist across the eight teacher models we test (see Appendix~\ref{app:behaviour-details} for details).

\subsection[Prompting and Behavioral Filtering]{Prompting and Behavioral Filtering}
\label{sec:mitigation}

Repeatedly generating and filtering trajectories to obtain verified correct solutions is difficult to scale at the research level, where reliable verification is itself nontrivial. We therefore opt for behavioral filtering, hypothesizing that substantive reasoning attempts can provide useful supervision even when they contain mathematical errors. To encourage such attempts, we introduce a committed-attempt prompt that requires models to engage with the original problem and explicitly identify unresolved steps (Appendix~\ref{app:response-generation}). We sample $500$ questions from \dataset{} and generate two responses to each with \texttt{Qwen3-30B-A3B-Thinking}, one with the prompt and one without it. We use \texttt{Claude-Opus-5} to label each response for non-attempts and fabricated citations. A \emph{non-attempt} declares the problem open, refuses, or surveys known results instead of attempting the problem. \emph{Fabricated citations} refer to cited references that appear not to exist. The prompt reduces the non-attempt rate from $22\%$ to $0\%$ and the proportion of responses containing fabricated citations from $38\%$ to $7\%$. We then screen trajectories using curated keyword lists for phrases associated with abandonment, unsupported claims, and citations (Appendix~\ref{app:keyword-groups}).

\section[Experiment Setup]{Experiment Setup}
\label{sec:learning-setup}

\paragraph{Training Setup.}\label{sec:training-setup}~Our main experiments use full-parameter SFT across three instruction-tuned models, \texttt{SmolLM3-3B}~\citep{bakouch2025smollm3}, \texttt{Qwen3-4B-Instruct-2507}~\citep{yang2025qwen3}, and \texttt{Olmo-3-7B-Instruct}~\citep{olmo2025olmo}. For SmolLM3, we use the \texttt{IT-LC-Expert} checkpoint as the SFT baseline. We fine-tune each model separately on \textsc{ResearchMath} ($77$K examples) and two baseline datasets, \texttt{Nemotron-SFT-Math-v4} ($86$K)~\citep{nvidia2026nemotronsftmathv4} and DASD ($50$K)~\citep{yan2026distribution}. Each training set contains approximately $0.9$B tokens (Table~\ref{tab:main-results}). We also compare DASD $100$K with a mixture of DASD $50$K and \textsc{ResearchMath} $77$K, both matched to approximately $2$B tokens (Figure~\ref{fig:training-trajectory-panels}). Appendix~\ref{app:training} provides further details.

\paragraph{Evaluation.}\label{sec:evaluation-setup}~For evaluation, we consider olympiad-level benchmarks (AIME 2024--2026 and SOOHAK Mini) and graduate- and research-level benchmarks (HLE, SOOHAK Challenge, and \texttt{ArxivMath}~\citep{dekoninck2026beyond}). We filter HLE and SOOHAK to include questions with integers only. \texttt{ArxivMath} spans December 2025 through April 2026 and includes integer and short-form symbolic answers. We use \texttt{math-verify}\footnote{\url{https://github.com/huggingface/Math-Verify}} for scoring. Finally, to evaluate proof-style reasoning, we use the development set of \texttt{QEDBench}~\citep{gonzalez2026qedbench}, with \texttt{Gemini-3.1-Pro} as the judge. All scores use a $0$ to $100$ scale, with QEDBench's mean multiplied by $100$.

\section{Main Results}
\label{sec:learning-results}


\paragraph{Unverified Research-Level Attempts Improve Downstream Performance.}~In Table~\ref{tab:main-results}, we train three models on \textsc{ResearchMath} for six epochs, exposing each model to approximately $6$B tokens of unverified research-level attempts. Averaged across the three model families and five benchmarks, \textsc{ResearchMath} improves performance by $4.0$ percentage points over the corresponding starting checkpoints. This exceeds the average gains from DASD ($+3.8$ points) and \texttt{Nemotron-SFT-Math-v4} ($+0.015$ points). We also find that the gains from each dataset differ substantially. DASD gains $9.5$ points on olympiad-level benchmarks but essentially none on graduate- and research-level benchmarks ($+0.0$ points), relative to the starting checkpoints. \textsc{ResearchMath} improves the two groups by $6.9$ and $2.1$ points, respectively, versus $+0.8$ and $-0.5$ for \texttt{Nemotron-SFT-Math-v4}.

\begin{table*}[h]
\centering
\scriptsize
\setlength{\tabcolsep}{2.5pt}
\caption{\footnotesize \textbf{Full-parameter SFT results across model families and mathematical difficulty levels.} We evaluate olympiad-level performance on AIME 2024--2026 and SOOHAK Mini, and graduate- and research-level performance on HLE, SOOHAK Challenge, and ArxivMath. Group averages are unweighted arithmetic means across the corresponding benchmarks. Benchmark scores are means with standard deviations over three runs in a stochastic setting. Bold marks the best result within each model block and column. SH. abbreviates SOOHAK.}
\label{tab:main-results}
\begin{tabular*}{\textwidth}{@{\extracolsep{\fill}}lccccccc@{}}
\toprule
\multicolumn{1}{c}{\raisebox{-1.7ex}[0pt][0pt]{Model}} & \multicolumn{3}{c}{Olympiad Level} & \multicolumn{4}{c}{Graduate/Research} \\
\cmidrule(lr){2-4}\cmidrule(lr){5-8}
 & AIME 24-26 & SH. Mini & Avg. & HLE & SH. Challenge & ArxivMath & Avg. \\
\midrule
SmolLM3-3B (SFT baseline) & $8.15{\scriptstyle\,\pm\,1.3}$ & $26.0{\scriptstyle\,\pm\,1.9}$ & $17.08$ & $6.35{\scriptstyle\,\pm\,0.8}$ & $1.7{\scriptstyle\,\pm\,1.3}$ & $6.53{\scriptstyle\,\pm\,0.4}$ & $4.86$ \\
\quad + Nemotron-SFT-Math-v4 & $12.22{\scriptstyle\,\pm\,2.9}$ & $26.6{\scriptstyle\,\pm\,0.4}$ & $19.41$ & $7.41{\scriptstyle\,\pm\,0.5}$ & $2.1{\scriptstyle\,\pm\,0.9}$ & $5.59{\scriptstyle\,\pm\,0.7}$ & $5.03$ \\
\quad + DASD & $\mathbf{22.22}{\scriptstyle\,\mathbf{\pm\,2.9}}$ & $32.1{\scriptstyle\,\pm\,2.7}$ & $\mathbf{27.16}$ & $6.35{\scriptstyle\,\pm\,1.4}$ & $2.5{\scriptstyle\,\pm\,0.4}$ & $4.43{\scriptstyle\,\pm\,0.8}$ & $4.43$ \\
\quad + ResearchMath (Ours) & $18.15{\scriptstyle\,\pm\,2.8}$ & $\mathbf{35.7}{\scriptstyle\,\mathbf{\pm\,3.1}}$ & $26.93$ & $\mathbf{8.89}{\scriptstyle\,\mathbf{\pm\,1.1}}$ & $\mathbf{3.5}{\scriptstyle\,\mathbf{\pm\,1.1}}$ & $\mathbf{7.93}{\scriptstyle\,\mathbf{\pm\,2.9}}$ & $\mathbf{6.77}$ \\
\midrule
Qwen3-4B-Instruct-2507 & $54.44{\scriptstyle\,\pm\,4.0}$ & $55.4{\scriptstyle\,\pm\,0.4}$ & $54.92$ & $9.31{\scriptstyle\,\pm\,1.9}$ & $4.1{\scriptstyle\,\pm\,0.6}$ & $13.05{\scriptstyle\,\pm\,1.5}$ & $8.82$ \\
\quad + Nemotron-SFT-Math-v4 & $52.96{\scriptstyle\,\pm\,5.5}$ & $50.4{\scriptstyle\,\pm\,0.6}$ & $51.68$ & $8.57{\scriptstyle\,\pm\,0.8}$ & $3.9{\scriptstyle\,\pm\,0.9}$ & $8.39{\scriptstyle\,\pm\,0.7}$ & $6.95$ \\
\quad + DASD & $\mathbf{64.81}{\scriptstyle\,\mathbf{\pm\,1.7}}$ & $54.6{\scriptstyle\,\pm\,0.4}$ & $\mathbf{59.71}$ & $9.21{\scriptstyle\,\pm\,1.5}$ & $4.3{\scriptstyle\,\pm\,0.4}$ & $9.79{\scriptstyle\,\pm\,2.1}$ & $7.77$ \\
\quad + ResearchMath (Ours) & $51.11{\scriptstyle\,\pm\,4.0}$ & $\mathbf{55.8}{\scriptstyle\,\mathbf{\pm\,0.2}}$ & $53.46$ & $\mathbf{10.58}{\scriptstyle\,\mathbf{\pm\,2.3}}$ & $\mathbf{5.0}{\scriptstyle\,\mathbf{\pm\,2.5}}$ & $\mathbf{16.08}{\scriptstyle\,\mathbf{\pm\,0.7}}$ & $\mathbf{10.55}$ \\
\midrule
Olmo-3-7B-Instruct & $45.19{\scriptstyle\,\pm\,2.3}$ & $49.1{\scriptstyle\,\pm\,1.1}$ & $47.15$ & $6.98{\scriptstyle\,\pm\,1.1}$ & $2.7{\scriptstyle\,\pm\,0.9}$ & $8.39{\scriptstyle\,\pm\,1.4}$ & $6.02$ \\
\quad + Nemotron-SFT-Math-v4 & $51.11{\scriptstyle\,\pm\,4.8}$ & $49.7{\scriptstyle\,\pm\,2.8}$ & $50.41$ & $7.30{\scriptstyle\,\pm\,0.6}$ & $\mathbf{3.9}{\scriptstyle\,\mathbf{\pm\,1.3}}$ & $7.46{\scriptstyle\,\pm\,2.3}$ & $6.22$ \\
\quad + DASD & $\mathbf{67.04}{\scriptstyle\,\mathbf{\pm\,2.6}}$ & $54.4{\scriptstyle\,\pm\,1.8}$ & $\mathbf{60.72}$ & $8.99{\scriptstyle\,\pm\,2.4}$ & $2.9{\scriptstyle\,\pm\,1.6}$ & $10.72{\scriptstyle\,\pm\,0.4}$ & $7.54$ \\
\quad + ResearchMath (Ours) & $62.59{\scriptstyle\,\pm\,3.6}$ & $\mathbf{56.1}{\scriptstyle\,\mathbf{\pm\,1.2}}$ & $59.35$ & $\mathbf{10.26}{\scriptstyle\,\mathbf{\pm\,1.6}}$ & $3.5{\scriptstyle\,\pm\,1.8}$ & $\mathbf{11.89}{\scriptstyle\,\mathbf{\pm\,3.1}}$ & $\mathbf{8.55}$ \\
\bottomrule
\end{tabular*}
\end{table*}

\begin{figure*}[t]
  \centering
  \includegraphics[width=\textwidth]{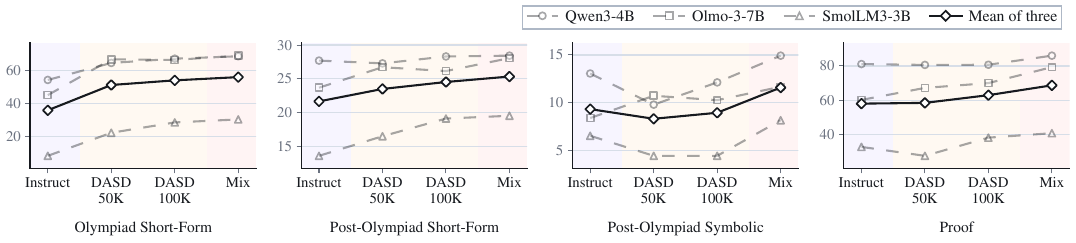}
  \caption{\footnotesize \textbf{Full-parameter SFT results comparing DASD-only scaling with mixed DASD and ResearchMath supervision.} From left to right, the panels evaluate AIME 2024--2026 (olympiad short-form), HLE and SOOHAK (graduate- and research-level short-form), \texttt{ArxivMath} (graduate- and research-level symbolic), and \texttt{QEDBench} (proof). Gray lines denote individual models, and black denotes the three-model mean.}
  \label{fig:training-trajectory-panels}
  \vspace{-4pt}
  \end{figure*}
  
\paragraph{Research-Level Trajectories Complement Existing Mathematical Post-Training Data.}~Another key question is whether \textsc{ResearchMath} provides a benefit orthogonal to that of existing mathematical post-training data. We test whether this advantage translates into complementarity when the datasets are combined. If its training signal were redundant, mixing it with DASD should offer no advantage over scaling DASD alone. In Figure~\ref{fig:training-trajectory-panels}, we compare DASD $100$K with a token-matched mixture of DASD $50$K and \textsc{ResearchMath} $77$K. The mixture improves all three models across all four task families, with average gains of $2.0$ points on olympiad short-form tasks, $0.8$ and $2.6$ on graduate- and research-level short-form and symbolic tasks, respectively, and $5.7$ on QEDBench. Notably, research-level trajectories not only improve performance on graduate- and research-level and proof tasks, but also strengthen olympiad-level performance when combined with DASD.

\section[Why Do Unverified Research-Level Trajectories Help?]{Why Do Unverified Research-Level Trajectories Help?}
\label{sec:correctness-diversity}

The gains in Section~\ref{sec:learning-results} raise a question about what student models learn from \textsc{ResearchMath}, given that only a small fraction of its teacher trajectories are judged correct (Section~\ref{sec:trajectory-correctness}). We hypothesize that even an unsuccessful attempt may expose a student model to mathematical concepts and reasoning approaches that are uncommon in existing training data, thereby helping it generalize to a broader range of mathematical problems. To examine this possibility, we conduct two complementary comparisons. First, we relate downstream gains to the mathematical content of the three corpora, considering differences in trajectory length and teacher model (Section~\ref{sec:research-level-content}). We then compare reasoning diversity across traces matched by thought-unit count, asking whether \textsc{ResearchMath} provides a broader range of reasoning patterns (Section~\ref{sec:reasoning-diversity}).

\subsection{Research-Level Content Contributes to Downstream Gains}
\label{sec:research-level-content}

\begin{table}[t]
\centering
\fontsize{8}{9.5}\selectfont
\setlength{\tabcolsep}{3pt}
\caption{\footnotesize \textbf{Training data characteristics and downstream gains.} Length is the mean number of reasoning tokens per trajectory. Gains are point differences from the starting checkpoints, averaged across the three model families and the benchmarks in each group (Table~\ref{tab:main-results}). Math SE/MO abbreviates Math StackExchange/MathOverflow.}
\label{tab:training-data-comparison}
\begin{tabular*}{\textwidth}{@{\extracolsep{\fill}}lcp{0.18\textwidth}lcc@{}}
\toprule
Dataset & Length & Teacher models & Problem sources & Olympiad gain & Graduate/Research \\
\midrule
\texttt{Nemotron-SFT-Math-v4} & $10$K & \texttt{DeepSeek-V4-Pro} & AoPS + Math SE/MO & $+0.8$ & $-0.5$ \\
DASD & $13$K & \texttt{GPT-OSS-120B} & AceReason & $+9.5$ & $+0.0$ \\
\textsc{ResearchMath} & $11$K & \texttt{Qwen3-30B-A3B} & Research Papers & $+6.9$ & $+2.1$ \\
\bottomrule
\end{tabular*}
\end{table}

What distinguishes \textsc{ResearchMath} from the baseline datasets? In Table~\ref{tab:training-data-comparison}, we compare their trajectory lengths, teacher models, and problem sources. First, longer trajectories do not consistently yield larger average gains, as \textsc{ResearchMath} outperforms DASD despite its trajectories being shorter. \textsc{ResearchMath} also outperforms \texttt{Nemotron-SFT-Math-v4} despite the latter's larger teacher, suggesting that teacher size alone does not explain the gains. A notable distinction lies in the sources of the training problems. DASD draws its math questions primarily from \texttt{AceReason}~\citep{liu2025acereason}, whose sources include \texttt{NuminaMath-CoT}~\citep{numina_math_datasets} and \texttt{OpenMathReasoning}~\citep{moshkov2025aimo2}. \texttt{Nemotron-SFT-Math-v4} draws from AoPS\footnote{\url{https://artofproblemsolving.com/}} and Math StackExchange/MathOverflow and removes proof-style questions~\citep{nvidia2026nemotronsftmathv4}. Both controls therefore emphasize competition-style or answer-verifiable problems, whereas \textsc{ResearchMath} is built from researcher-authored open problems. Consistent with these source distributions, DASD leads on olympiad-level benchmarks and \textsc{ResearchMath} on graduate- and research-level benchmarks. \textbf{Together, these comparisons suggest that the mathematical level of supervision shapes where gains emerge, with research-level content offering benefits beyond olympiad-focused data.}

\subsection[Reasoning Diversity May Provide Complementary Supervision]{Reasoning Diversity May Provide Complementary Supervision}
\label{sec:reasoning-diversity}

\begin{table}[!ht]
\centering
\fontsize{8}{9.5}\selectfont
\setlength{\tabcolsep}{3pt}
\caption{\footnotesize \textbf{Reasoning diversity.} From $4{,}000$ sampled trajectories per dataset, we retain $3{,}893$ traces matched by thought-unit count. Values are mean $\pm$ SD over ten $2{,}000$-trace subsamples.}
\label{tab:trajectory-diversity-correctness}
\begin{tabular*}{\textwidth}{@{\extracolsep{\fill}}lccc@{}}
\toprule
Dataset & Vendi & Effective dimensionality & Reasoning-to-question Vendi ratio \\
\midrule
\texttt{Nemotron-SFT-Math-v4} $86$K & $101.3\pm0.9$ & $96.4\pm1.0$ & $0.672\pm0.003$ \\
DASD $50$K & $117.0\pm2.1$ & $103.3\pm1.4$ & $0.707\pm0.003$ \\
\textsc{ResearchMath} $77$K & $\mathbf{139.4\pm1.2}$ & $\mathbf{132.2\pm1.7}$ & $\mathbf{1.051\pm0.004}$ \\
\bottomrule
\end{tabular*}
\end{table}

Another possible explanation for these gains is the diversity of reasoning content in research-level trajectories. To examine this possibility, we compare trajectory diversity across the three datasets. Starting with $4{,}000$ sampled trajectories per dataset, we control for length by matching thought-unit counts from \textsc{ThinkProbe}~\citep{kerkouri2026thinkprobe}. This yields $3{,}893$ trajectories per dataset. Each trajectory and its question are embedded separately with \texttt{Qwen3-Embedding-8B}~\citep{zhang2025qwen3}. With $n$ unit-normalized embeddings as rows of $X$, let $\lambda_i$ be the eigenvalues of the normalized cosine kernel $XX^\top/n$ and $\mu_i$ those of the centered covariance. Our three metrics are
\[
V(X)=\exp\!\left(-\sum_i\lambda_i\log\lambda_i\right),\qquad
D_{\mathrm{eff}}(X)=\frac{(\sum_i\mu_i)^2}{\sum_i\mu_i^2},\qquad
R/Q=\frac{V(X_r)}{V(X_q)}.
\]
\emph{Vendi score} $V$~\citep{friedman2022vendi} summarizes embedding similarities as an effective number of distinct items. \emph{Effective dimensionality} $D_{\mathrm{eff}}$ examines variation around each dataset's mean embedding and emphasizes larger variance directions. The \emph{reasoning-to-question Vendi ratio} $R/Q$ compares the diversity of matched reasoning embeddings $X_r$ and question embeddings $X_q$. Values above one indicate greater embedding diversity in the reasoning traces than in their questions.

As shown in Table~\ref{tab:trajectory-diversity-correctness}, \textsc{ResearchMath} achieves the highest Vendi score. It also leads in effective dimensionality, suggesting robustness across these two measures of embedding diversity. Moreover, \textsc{ResearchMath} is the only dataset with $R/Q>1$ ($1.051$, versus $0.707$ for DASD and $0.672$ for Nemotron). This may reflect both broader subject coverage than olympiad-focused data and more varied attempts elicited by research-level problems. Together, these observations suggest that \textsc{ResearchMath} contains more diverse reasoning content than the baseline datasets, providing a plausible explanation for its downstream benefits. See Appendix~\ref{app:correctness-diversity} for further details.

\section{Conclusion}
\label{sec:conclusion}
This work introduces \dataset{}, an agent-curated collection of $14{,}056$ research-level problems across $11$ fields from $1{,}233$ sources, alongside $220$K teacher trajectories. Under full-parameter training, \textsc{ResearchMath} yields a $4.0$-point average gain across three model families and five benchmarks, outperforming existing datasets. Moreover, \textsc{ResearchMath} yields larger gains on graduate- and research-level benchmarks and further improves performance when combined with DASD. These findings show that unverified research-level attempts can provide useful post-training supervision without establishing that models solve the underlying open problems. We release all resources to support future work on research-level mathematics.

\bibliography{custom}
\bibliographystyle{iclr2027_conference}
\clearpage
\appendix

\onecolumn
\lstset{
  literate=
    {⊂}{{$\subset$}}1
    {ρ}{{$\rho$}}1
    {∈}{{$\in$}}1
    {₁}{{\ensuremath{{}_1}}}1
    {₂}{{\ensuremath{{}_2}}}1
    {ℝ}{{$\mathbb{R}$}}1
    {≥}{{$\ge$}}1
    {–}{{\textendash}}1
}

\onecolumn
\section*{Appendix Contents}

\begingroup
\setlength{\parindent}{0pt}
\newcommand{\appcontentsline}[2]{%
  \hyperref[#1]{\ref*{#1}\quad #2}\dotfill\pageref*{#1}\par
}
\newcommand{\appsubcontentsline}[2]{%
  \hspace*{1.5em}\hyperref[#1]{\ref*{#1}\quad #2}\dotfill\pageref*{#1}\par
}

\appcontentsline{app:source-example-comparison}{Example Source Comparisons}
\appcontentsline{app:self-containment-audit}{Self-Containment Review}
\appcontentsline{app:human-audit}{Human Expert Review}
\appcontentsline{app:decontamination}{Decontamination Details}
\appcontentsline{app:benchmark-details}{Control Benchmark Details}
\appsubcontentsline{app:length-controlled-difficulty}{Length-Controlled Difficulty Comparison}
\appcontentsline{app:keyword-judge-metric}{Keyword and Judge Metric Details}
\appsubcontentsline{app:metric-descriptions}{Full Metric Descriptions}
\appsubcontentsline{app:keyword-groups}{Keyword Groups}
\appsubcontentsline{app:judge-annotation}{Reference Verification}
\appsubcontentsline{app:aggregation}{Aggregation}

\appcontentsline{app:filtering-details}{Near-Duplicates Filtering Details}
\appsubcontentsline{app:sim-distribution}{Pairwise Similarity Distribution}
\appsubcontentsline{app:gpt5.5-judges}{\texttt{GPT-5.5} Judgments Near Decision Boundary}

\appcontentsline{app:behaviour-details}{Reasoning Behavior Details}
\appsubcontentsline{app:eight-model-study}{Eight-Model Study on \dataset{}}
\appsubcontentsline{app:conjecture}{Our Conjecture: Why Do Newer Models Fabricate More?}
\appsubcontentsline{app:behavior-counter-rates}{Behavior-Counter Rates}
\appsubcontentsline{app:cross-benchmark}{Cross-Benchmark Patterns}
\appsubcontentsline{app:per-model-reference-verification}{Per-Model Reference Verification}

\appcontentsline{app:license}{License and Release}

\appcontentsline{app:training}{Training Details}

\appcontentsline{app:correctness-diversity}{Trajectory Correctness and Diversity}

\appcontentsline{app:prompts}{Prompts}
\appsubcontentsline{app:dataset-generation-agents}{Dataset Generation Agents}
\appsubcontentsline{app:difficulty-comparision}{Difficulty Comparison}
\appsubcontentsline{app:response-generation}{Response Generation}
\appsubcontentsline{app:factuality-metrics}{Factuality Metrics}
\appsubcontentsline{app:approach-correctness-prompt}{Approach-Correctness Judging}

\endgroup

\clearpage
\onecolumn

\section{Example Source Comparisons}
\label{app:source-example-comparison}

Tables~\ref{tab:source-example-comparison} and~\ref{tab:source-example-comparison-hodge} illustrate the source-level contrast discussed in Section~\ref{sec:collection-process}. The first pair stays within number theory/arithmetic geometry; the second pairs a broad algebraic-geometry grand challenge with a narrower modern question about hyperkähler Chow rings.

\begin{table*}[h]
\centering
\scriptsize
\setlength{\tabcolsep}{4pt}
\renewcommand{\arraystretch}{1.25}
\caption{\footnotesize Side-by-side example of a classical arithmetic-geometry grand-challenge source and a narrower contemporary open-problem source represented in \dataset{}.}
\label{tab:source-example-comparison}
\begin{tabular}{p{0.18\textwidth}p{0.38\textwidth}p{0.38\textwidth}}
\toprule
 & Grand-challenge entry & Contemporary workshop-style entry \\
\midrule
Dataset record & \texttt{google\_\_mppc/q\_001} & \texttt{07-workshop-problems/q\_007} \\
Source & \emph{The Millennium Prize Problems} & \emph{Some Open Problems About Diophantine Equations} \\
Area & Number theory; elliptic curves and $L$-functions & Number theory; rational points on hyperelliptic curves \\
Problem form & Prove the Birch--Swinnerton-Dyer conjecture: for an elliptic curve $C/\mathbb{Q}$, the order of vanishing of $L(C,s)$ at $s=1$ equals the Mordell--Weil rank of $C(\mathbb{Q})$. & Prove unconditionally that the genus-$2$ curve
$$y^2=-3x^6-x^5+2x^4+2x^2-3x-3$$
has no rational points over $\mathbb{Q}$, given that this is known assuming BSD for its Jacobian. \\
Status & Open & Unknown \\
Comment & This is the kind of classical, high-visibility challenge problem that appears in many public lists of unsolved mathematics. & This is narrower and more local: it sits near BSD and rational-points methods, but asks for a concrete unconditional proof for one curve. It better represents the workshop and survey questions that form much of \dataset{}. \\
Source & \href{https://www.claymath.org/library/monographs/MPPc.pdf}{Clay monograph PDF} & \href{https://pub.math.leidenuniv.nl/~evertsejh/07-workshop-problems.pdf}{Leiden workshop PDF} \\
\bottomrule
\end{tabular}
\end{table*}

\begin{table*}[h]
\centering
\scriptsize
\setlength{\tabcolsep}{4pt}
\renewcommand{\arraystretch}{1.25}
\caption{\footnotesize Second side-by-side example: a broad algebraic-geometry grand challenge compared with a narrower contemporary hyperkähler/Chow-ring problem represented in \dataset{}.}
\label{tab:source-example-comparison-hodge}
\begin{tabular}{p{0.18\textwidth}p{0.38\textwidth}p{0.38\textwidth}}
\toprule
 & Grand-challenge entry & Contemporary survey-style entry \\
\midrule
Dataset record & \texttt{google\_\_mppc/q\_004} & \texttt{1002.4321/q\_010} \\
Source & \emph{The Millennium Prize Problems} & arXiv problem list on compact hyperkähler manifolds \\
Area & Algebraic geometry; Hodge theory and algebraic cycles & Algebraic geometry; hyperkähler manifolds and Chow rings \\
Problem form & Prove the Hodge conjecture: every rational Hodge class on a smooth projective complex variety is a $\mathbb{Q}$-linear combination of cohomology classes of algebraic subvarieties. & For every projective hyperkähler manifold $X$, prove that the subring of $\mathrm{CH}(X)$ generated by $\mathrm{Pic}(X)$ and the Chern classes of $X$ injects into cohomology under the cycle-class map. \\
Status & Open & Partially solved \\
Comment & This is a universal conjecture about the relation between Hodge-theoretic and algebraic cycles across all smooth projective complex varieties. & This is much more local: it asks for a specific Chow-ring injectivity statement in the modern setting of hyperkähler geometry. It uses current objects and tools while still sitting near the Hodge/cycle-theoretic grand challenge. \\
Source & \href{https://www.claymath.org/library/monographs/MPPc.pdf}{Clay monograph PDF} & \href{https://arxiv.org/abs/1002.4321}{arXiv:1002.4321} \\
\bottomrule
\end{tabular}
\end{table*}

\section[Self-Containment Review]{Self-Containment Review}
\label{app:self-containment-audit}

To quantify whether refinement makes questions usable without the source document, we run a first-pass automatic review on $500$ randomly sampled released records. For each record, Codex with GPT-5.5 labels both the original extracted question and the refined standalone question as self-contained or not. A statement is counted as self-contained only if a mathematically trained reader can understand the task from the text alone, without source-local notation, missing definitions, or references to external sections, figures, or problem numbers. Table~\ref{tab:self-containment-validation} shows that original extracted questions are self-contained in $67.2\%$ of sampled cases, while refined standalone questions are self-contained in $94.2\%$. The refiner turns $142/500$ initially non-self-contained snippets into self-contained questions, leaving $29/500$ refined questions flagged for remaining context gaps. Table~\ref{tab:self-containment-examples} shows representative before/after examples, including a remaining failure case.

\begin{table}[t]
\centering
\scriptsize
\setlength{\tabcolsep}{3pt}
\caption{\footnotesize \textbf{Automatic self-containment review.} We sample $500$ released records and ask Codex (GPT-5.5) to label both the original extracted question and the refined standalone question as self-contained or not.}
\label{tab:self-containment-validation}
\begin{tabular}{@{}p{0.56\columnwidth}rr@{}}
\toprule
Review item & Count & Rate \\
\midrule
Original extracted question & $336/500$ & $67.2\%$ \\
Refined standalone question & $471/500$ & $94.2\%$ \\
\midrule
Original no, refined yes & $142/500$ & $28.4\%$ \\
Original yes, refined no & $7/500$ & $1.4\%$ \\
\bottomrule
\end{tabular}
\end{table}

\begin{table*}[t]
\centering
\scriptsize
\setlength{\tabcolsep}{4pt}
\renewcommand{\arraystretch}{1.25}
\caption{\footnotesize Examples from the self-containment review. The first two rows show originally non-self-contained extractions that become self-contained after refinement. The final row shows a remaining failure case in which both the original extraction and refined statement still depend on source-local definitions.}
\label{tab:self-containment-examples}
\begin{tabular}{@{}p{0.11\textwidth}p{0.30\textwidth}p{0.50\textwidth}@{}}
\toprule
Case & Original extraction judged not self-contained & Refined standalone problem \\
\midrule
Geodesics &
Suppose $\operatorname{dist}(\alpha(t), \beta(t)) \to 0$ as $t \to -\infty$. &
Let $(M,g)$ be a complete, simply connected Riemannian manifold without conjugate points. Let $\alpha,\beta:\mathbb{R}\to M$ be two distinct unit-speed geodesics, and denote the Riemannian distance by $\operatorname{dist}$. Suppose that $\operatorname{dist}(\alpha(t),\beta(t))\to 0$ as $t\to-\infty$. Does it follow that $\operatorname{dist}(\alpha(t),\beta(t))\to\infty$ as $t\to+\infty$? \\
\midrule
Words &
Let $\Sigma$ be an alphabet and let $w \in \Sigma^{*}$ be a $p$-free word. Then $w$ is a factor of a maximal $p$-free word over $\Sigma$. &
Let $\Sigma$ be a finite alphabet and let $\Sigma^{*}$ denote the set of finite words over $\Sigma$. A pattern is a non-empty word $p$ over an alphabet of variables. An occurrence of $p$ in a word $w\in\Sigma^{*}$ is a factor of $w$ of the form $h(p)$, where $h$ is a non-erasing morphism. A word $w$ is $p$-free if no factor of $w$ is an occurrence of $p$. A $p$-free word is maximal if it cannot be extended on the left or right while remaining $p$-free. Prove or disprove: for every alphabet $\Sigma$, every pattern $p$, and every $p$-free word $w\in\Sigma^{*}$, there exists a maximal $p$-free word over $\Sigma$ that contains $w$ as a factor. \\
\midrule
Remaining gap &
Fix $p=c/n$ with $c>0$. Is it true that, with high probability for $G_{n,p}$, the width $s(G_{n,p})$ of the independence complex satisfies $s(G_{n,p})=(1+o(1)) \max_r |Q(G_{n,p})^{(r)}|$? &
Let $G$ be a finite simple graph with independence complex $I(G)$, whose faces are the independent sets of $G$. Following Meshulam's recursive framework, one associates to $G$ a nonnegative integer parameter $s(G)$ and a finite sequence $Q(G)=(Q(G)^{(0)},Q(G)^{(1)},\ldots)$ produced by the standard recursive process used to bound the topological connectivity of $I(G)$. For fixed $c>0$ and $G_{n,p}$ with $p=c/n$, is it true with high probability that $s(G_{n,p})=(1+o(1))\max_r |Q(G_{n,p})^{(r)}|$? The refined version still depends on the source for the exact recursive definition of $Q(G)$ and $s(G)$. \\
\bottomrule
\end{tabular}
\end{table*}

\section[Human Expert Review]{Human Expert Review}
\label{app:human-audit}

This appendix documents the protocol and qualitative evidence for the human expert review summarized in Section~\ref{sec:human-audit}. The review asks a deliberately narrow question. Given only a rewritten statement, can a specialist understand the problem and begin working on it? It does not rederive the mathematics, verify open or solved status, or evaluate the taxonomy. Its purpose is to detect missing hypotheses, definitions, quantifiers, and other context that automated checks may overlook.

\paragraph{Annotators and Sampling.}~We recruited four independent mathematics PhD students through an open call. They were not involved in building \dataset{}. Their specialties were number theory and arithmetic geometry ($99$ reviews), algebraic geometry ($13$), low-dimensional and quantum topology together with homogeneous dynamics and character varieties ($21$), and geometric topology, including convex projective geometry, Coxeter groups, and discrete subgroups of Lie groups ($5$). Each expert received a field-matched pool of roughly $1{,}000$ problems and selected items they felt familiar enough to review. The experts were asked to complete at least $25$ reviews, although three joined later and returned fewer. They were compensated at \$26 per review. Each review contained five ratings, a free-text error note or an explicit statement that no error was found, a written opinion, and an optional suggested fix. The fix was optional because some defects cannot be repaired confidently without consulting the source or the problem author.

\paragraph{Reference Reader and Rubric.}~Experts were guided to judge each statement from the perspective of a specialist in the relevant field who may consult standard background literature but does not have the source paper. Q1 (\emph{notation}) asks whether source-specific and nonstandard symbols are defined. Q2 (\emph{terminology}) asks whether every required concept is standard and findable or defined in the statement. Q3 (\emph{assumptions and well-posedness}) asks whether the hypotheses needed to make the problem well-posed are explicit. Q4 (\emph{goal clarity}) asks whether the task to prove, compute, construct, or decide is verifiable. Q5 (\emph{no dangling references}) asks whether the statement avoids pointers to absent figures, equations, or results. The nominal scale runs from $1$ to $5$, with $5$ indicating that the dimension is fully satisfied and $1$ indicating that the missing information prevents the reference reader from proceeding. In practice, the experts used only $1$, $3$, and $5$, so the instrument functions as a three-point scale.

\paragraph{Completed Sample.}~The survey contains $141$ submissions. We normalize reviewer name and email variants and treat repeated submissions by the same reviewer for the same problem identifier as revisions, retaining the latest timestamp. Removing three earlier submissions yields $138$ reviews associated with $137$ distinct problem identifiers. One identifier occurs in reviews by two experts, but the corresponding problem descriptions differ, as discussed below. Table~\ref{tab:human-expert-audit} reports unweighted means and top-rating proportions over the retained reviews. Of these, $115$ ($83.3\%$) give the top rating on all five dimensions and $4$ ($2.9\%$) give a minimum rating on at least one dimension.

\paragraph{Error Notes and Revisions.}~Experts provided error or revision notes in $19$ retained reviews ($13.8\%$). Responses stating that no error was found, including the marker ``x,'' are excluded from this count. The notes combine mathematical concerns, ambiguities, and optional editorial improvements, so their frequency should not be interpreted as a verified mathematical error rate.

\paragraph{Specific Defect Reports.}~We classify a note as a specific defect report when it identifies a concrete error in the mathematical statement or task instruction that could mislead or block the reference reader. Five of the $19$ notes meet this criterion ($26.3\%$ of notes and $3.6\%$ of all $138$ reviews). The count concerns specific errors reported by reviewers. General reservations about the problem or its difficulty are excluded. The following English summaries describe the five reports.

\begin{enumerate}[leftmargin=1.6em, itemsep=2pt, topsep=3pt]
\item \textbf{Incorrect description of a mathematical object.} A reviewer identified a mismatch between $G(\mathbb{C})$ and its description as an associated compact real Lie group. The reviewer proposed ``complex Lie group'' as the intended wording.
\item \textbf{Equivalence requiring an unstated assumption.} A reviewer flagged a parenthetical characterization using the ring of regular functions as requiring an affine-variety assumption. The suggested correction was to remove the equivalence or state the additional assumption.
\item \textbf{Invalid equivalence.} A reviewer stated that two conditions presented as equivalent were not equivalent. If the intended task was to find a sequence making them equivalent, the reviewer recommended stating that task explicitly.
\item \textbf{Truncated instruction.} A reviewer reported that a prompt contained only ``or disprove'' where the intended instruction was ``prove or disprove.''
\item \textbf{Condition affecting the problem's status.} A reviewer changed a strict pinching condition to a weak one and explained that the interpretation affects whether the negatively curved manifold question is already resolved. This revised submission replaces the reviewer's earlier response.
\end{enumerate}

\paragraph{Other Revision Notes.}~The remaining $14$ notes mainly offer suggestions for improving notation, presentation, and clarity. Suggestions include using clearer notation or LaTeX formatting, specifying field characteristic or the measure used for Lyapunov exponents, supplying standard definitions or an explicit matrix formula, adding a dimension condition to focus on a nontrivial case, and removing redundant exposition. General questions about the formulation are retained as feedback but are not counted as specific defect reports.

\paragraph{Examples of Positive Opinions.}~The following excerpts, translated from Korean, illustrate favorable overall assessments within the same set of $19$ reviews with error or revision notes. They concern the interest, scope, or usefulness of the problems and do not imply that the accompanying notes contain no substantive concerns.

\begin{enumerate}[leftmargin=1.6em, itemsep=2pt, topsep=3pt]
\item \textbf{Algebraic versus analytic isomorphism.} ``The problem is well-posed and has an appropriate scope to interest researchers. It is suitable as evaluation data.''
\item \textbf{Closed orbits in homogeneous dynamics.} ``This is a problem that could interest many mathematicians, not just those working in homogeneous dynamics.''
\item \textbf{Lyapunov exponents.} ``It could attract broad interest. Since it asks for an example, I think it is approachable through AI-assisted research.''
\end{enumerate}

\paragraph{Written Opinions and Agreement.}~Every retained review includes a written opinion. After trimming surrounding whitespace, these responses average approximately $182$ characters and range from $94$ to $468$. They discuss mathematical interest, difficulty, and suitability for evaluation, which the rating rubric does not quantify. Two experts used the same problem identifier and both assigned top ratings on all five dimensions, but one opinion discusses a length-spectrum question and the other an injectivity-radius bound. We retain both records as separate reviews, but the inconsistent descriptions leave their problem identity unverified. We therefore do not treat their matching ratings as evidence of agreement or estimate inter-annotator reliability from this sample.

\paragraph{Scope and Limitations.}~The $138$ retained reviews are associated with $137$ distinct problem identifiers in \dataset{}. They support conclusions about the reviewed sample, not a guarantee for all $14{,}056$ items. The review covers four specialties and does not establish quality in every domain. Experts self-selected problems from field-matched pools, so the sample may favor familiar, tractable, or interesting items. It remains highly imbalanced, with one expert contributing $99$ of the $138$ reviews ($71.7\%$). The effective three-point scale limits fine comparisons among mean ratings, and inter-annotator agreement is unmeasured. Consolidating repeat submissions depends on the recorded problem identifiers, whose cross-reviewer overlap contains the discrepancy noted above. The rubric evaluates statement self-containedness, not source fidelity, solvability, solution correctness, or research-level difficulty. The review was diagnostic, and flagged statements were not corrected in the released corpus. Subject to these limitations, the ratings support the self-containedness of most reviewed statements, while the written feedback identifies concerns for further checking.

\section[Decontamination Details]{Decontamination Details}
\label{app:decontamination}

We check for near-verbatim overlap between \dataset{} and every evaluation benchmark used in this paper. Each dataset record contributes both \texttt{original\_question} and \texttt{self\_contained\_problem}, giving $28{,}112$ indexed documents. We normalize each text to lowercase alphanumeric words, hash every word $n$-gram, build an inverted index over the benchmark queries, and stream each indexed document once. For each query, maximum containment is the number of shared $n$-grams divided by the total number of query $n$-grams. We run the analysis at $n=14$ and $n=18$. The benchmark suite contains \texttt{AIME} ($90$ problems), \texttt{HLE} ($315$), the English version of \texttt{SOOHAK} ($501$), \texttt{QEDBench} ($48$), and $143$ \texttt{ArxivMath} problems collected from December 2025 through April 2026. The five monthly \texttt{ArxivMath} splits contain $17$, $23$, $32$, $30$, and $41$ problems, respectively. No evaluable benchmark query shares even a single $14$-gram or $18$-gram with either \dataset{} field. The maximum containment is therefore $0$ at both values of $n$, and no query exceeds $0.10$. As a positive control, we insert $71$ verbatim \dataset{} problems into the query set. All $71$ controls are recovered with containment $1.0$ at both values of $n$, confirming that the index detects overlap when it is present. We mark $19$ queries at $n=14$ and $51$ at $n=18$ as unevaluable because they contain fewer than $n$ normalized words, rather than assigning them a containment score of $0$. This exact-match analysis detects near-verbatim reuse but not paraphrased or translated restatements.

\Needspace{28\baselineskip}
\section[Control Benchmark Details]{Control Benchmark Details}
\label{app:benchmark-details}

This appendix details the benchmark selection for the behavioral study in Appendix~\ref{app:eight-model-study}. \dataset{} has two defining properties: problems are \emph{research-level}, and their surface form is \emph{AI-refined} from a source paper. We choose four control benchmarks to isolate each property. To control for any artifact of the AI-refining step, we use \texttt{SOOHAK}~\citep{son2026soohak} and \texttt{Leipzig Tier-4}~\citep{sciencebench_leipzig_2026}, both research-level but human-authored. To study the effect of difficulty, we use the math subset of \texttt{HLE-Verified} (a version of Humanity's Last Exam~\citep{phan2025humanity} verified by \citet{zhai2026hle}) and \texttt{AIME}~\citep{aime24,aime25,aime26}. Both are easier than the research-level sets, with \texttt{AIME} being easiest. \texttt{AIME} combines questions from 2024, 2025, and 2026 for $90$ problems in total. We sample $90$ items from each of the other four benchmarks, with \texttt{SOOHAK} restricted to items labeled \emph{graduate} or beyond from the challenge subset, and all benchmarks further filtered to short-form-answer questions; this leaves \texttt{SOOHAK} with $86$ items, for $446$ prompts overall.

\subsection[Length-Controlled Difficulty Comparison]{Length-Controlled Difficulty Comparison}
\label{app:length-controlled-difficulty}

We complement the pairwise difficulty judgments in Section~\ref{sec:dataset-statistics} with a comparison of solver completion lengths that does not require an LLM judge. To control for differences in problem-statement length, we match the statement-length distributions of \dataset{}, \texttt{HLE-Verified}, \texttt{AIME}, \texttt{AceMath}, and \texttt{NuminaMath} using coarsened exact matching in $25$-token strata under the \texttt{cl100k} tokenizer. This yields $86$ problems per dataset ($430$ total), with a median statement length of $100$ tokens for every dataset.

We generate solutions with two independent solvers, \texttt{DeepSeek-V4-Pro} and \texttt{GLM-5.2}, using high reasoning effort, temperature $0.6$, and a $128$K-token budget. We measure the full completion length, including the reasoning trace and final answer, as a behavioral proxy for difficulty. Table~\ref{tab:length-control} reports mean completion lengths and ratios relative to \texttt{NuminaMath}.

On \dataset{}, the mean completion lengths are $21{,}439$ tokens for \texttt{DeepSeek-V4-Pro} and $27{,}747$ tokens for \texttt{GLM-5.2}, respectively $16.6$ and $17.2$ times those on \texttt{NuminaMath}. Both solvers also produce substantially longer completions on \dataset{} than on \texttt{AIME} and \texttt{AceMath}. Completion lengths are comparable to those on \texttt{HLE-Verified}, although their relative ordering depends on the solver. These results support the high difficulty of \dataset{} after controlling for problem-statement length.

\begin{table}[h]
\centering
\scriptsize
\setlength{\tabcolsep}{3pt}
\caption{\footnotesize \textbf{Length-controlled difficulty comparison.} Mean completion length (reasoning trace plus final answer) on $86$ statement-length-matched problems per dataset ($430$ total; median statement length $100$ tokens for every dataset), with ratios relative to \texttt{NuminaMath}. Solutions are generated at high reasoning effort with temperature $0.6$ and a $128$K-token budget. Bold marks the highest value per solver.}
\label{tab:length-control}
\begin{tabular}{@{}lrrrr@{}}
\toprule
 & \multicolumn{2}{c}{\texttt{DeepSeek-V4-Pro}} & \multicolumn{2}{c}{\texttt{GLM-5.2}} \\
\cmidrule(lr){2-3}\cmidrule(lr){4-5}
Dataset & Tokens & Ratio & Tokens & Ratio \\
\midrule
\dataset{} & $21{,}439$ & $16.6\times$ & $\mathbf{27{,}747}$ & $17.2\times$ \\
HLE-Verified & $\mathbf{32{,}778}$ & $25.4\times$ & $22{,}779$ & $14.1\times$ \\
AIME & $7{,}233$ & $5.6\times$ & $6{,}200$ & $3.8\times$ \\
AceMath & $2{,}026$ & $1.6\times$ & $2{,}468$ & $1.5\times$ \\
NuminaMath & $1{,}292$ & $1.0\times$ & $1{,}615$ & $1.0\times$ \\
\bottomrule
\end{tabular}
\end{table}

\section{Keyword and Judge Metric Details}
\label{app:keyword-judge-metric}
\label{app:keyword-metrics}

This appendix specifies the surface-form counters used in Appendix~\ref{app:eight-model-study}. All keyword matching is performed after lowercasing the analyzed text. A keyword group count is the sum of exact substring occurrences of all phrases in that group. The counters are descriptive trace features; they are not used as a standalone hallucination classifier.

\subsection{Full Metric Descriptions}
\label{app:metric-descriptions}

\paragraph{Rule-Based Counting.}~We use three curated phrase lists, each targeting a distinct phrasing pattern. The lists were assembled by the authors after reviewing dozens of model reasoning traces and collecting recurrent phrases that fit each pattern, and matching is performed against the lowercased trace (full lists in Appendix~\ref{app:keyword-groups}). \texttt{cite} matches citation-like nouns (e.g.\ ``paper''). \texttt{abandon} catches abandonment (e.g.\ ``cannot solve'', ``educated guess''). \texttt{assume} catches claims made without justification (e.g.\ ``known result'', ``i remember''). Two of these (\texttt{abandon}, \texttt{assume}) measure behavior, while \texttt{cite} measures factuality and bridges into the agent-judge below. Each counter increments by one per match, and per benchmark we report the \emph{row-hit rate} $\frac{1}{|T|}\sum_{i \in T}\mathbf{1}[n_{i,c} > 0]$, the fraction of traces in which counter $c$ matches at least once ($n_{i,c}$ is the match count in trace $i$ and $T$ is the set of traces). These rules are transparent, cheap, and chosen to broadly cover recurring failure patterns. Counting alone, however, cannot judge whether a given match is a real failure in context.

\paragraph{Agent-Judge.}~We inspect whether reference-like spans in the trace correspond to real sources. Because running an agent over a full reasoning trace is expensive, we use a two-stage pipeline. We slice each trace into newline-delimited blocks and use \texttt{GPT-5.4-nano} to review each block and extract reference-like spans (books, papers, website URLs). A search-enabled \texttt{Codex} agent then iterates over each span to confirm whether the span is genuine reference text (filtering out e.g.\ named mathematical theorems) and whether the referenced source exists on the web. We provide the surrounding block for reference, and require multiple web searches before every judgment. Prompts for the factuality pipeline are in Appendix~\ref{app:factuality-metrics}. This check does not establish mathematical correctness.

\subsection{Keyword Groups}
\label{app:keyword-groups}

\paragraph{Training-set construction.}~We construct the $77$K training set from the $220$K teacher-trajectory collection. All trajectories in this collection are generated using the committed-attempt prompt in Appendix~\ref{app:response-generation}. We first remove overly long examples and then apply keyword filtering using the phrase lists below. Finally, we randomly sample $77$K trajectories from the remaining pool for training.

\paragraph{\texttt{abandon}.}
This counter marks claims of being stuck, time-limited, or unable to complete the solution. The keyword list is:
\texttt{lack of progress},
\texttt{given the time},
\texttt{time constraints},
\texttt{too complex},
\texttt{not practical manually},
\texttt{i'm stuck},
\texttt{i am stuck},
\texttt{dead end},
\texttt{can't solve},
\texttt{cannot solve},
\texttt{without progress},
\texttt{hazard a guess},
\texttt{educated guess}.

\paragraph{\texttt{cite}.}
This counter marks references to source objects, external databases, or citation-like artifacts. The keyword list is:
\texttt{paper},
\texttt{book},
\texttt{article},
\texttt{textbook},
\texttt{monograph},
\texttt{survey},
\texttt{journal},
\texttt{proceedings},
\texttt{publication},
\texttt{arxiv},
\texttt{doi},
\texttt{wikipedia},
\texttt{mathworld},
\texttt{oeis},
\texttt{stackexchange},
\texttt{aops},
\texttt{art of problem solving},
\texttt{website},
\texttt{webpage},
\texttt{online source}.

\paragraph{\texttt{assume}.}
This counter combines assertive shortcuts and remembered-result language, both of which substitute confident assertion for derivation in the trace. The keyword list is:
\texttt{it can be shown},
\texttt{one can show},
\texttt{it is easy to see},
\texttt{clearly},
\texttt{obviously},
\texttt{intuitively},
\texttt{by symmetry},
\texttt{must be},
\texttt{should be},
\texttt{known result},
\texttt{standard result},
\texttt{i remember},
\texttt{similar problem online},
\texttt{look it up mentally},
\texttt{the problem implies},
\texttt{strongly suggests},
\texttt{well-known},
\texttt{it is known},
\texttt{i recall}.

\subsection[Reference Verification]{Reference Verification}
\label{app:judge-annotation}

\paragraph{Agent-Judge reference verification.}
This annotation is applied to traces that mention citation-like references, including papers, books, articles, arXiv identifiers, DOI-like strings, named sources, or database references. A Codex-based search agent queries the internet for the mentioned reference and records whether the referenced source appears to exist. The purpose is to separate genuine provenance signals from hallucinated bibliographic support. This annotation does not judge whether the source proves the model's claim; it only checks whether the cited object itself can be found.


\subsection{Aggregation}
\label{app:aggregation}

For each keyword group and Agent-Judge verification result, we aggregate by model, model family, and benchmark. The row-hit rate treats each trace as a binary hit for a counter. For reference verification, we report the fraction of traces containing at least one reference judged fake and the fake share among reference-like mentions. The benchmark trend view reports the average newer-minus-older delta across the DeepSeek, Kimi, and Qwen comparison pairs.

\begin{center}
\IfFileExists{figures/similarity_distribution.pdf}{\includegraphics[width=\textwidth]{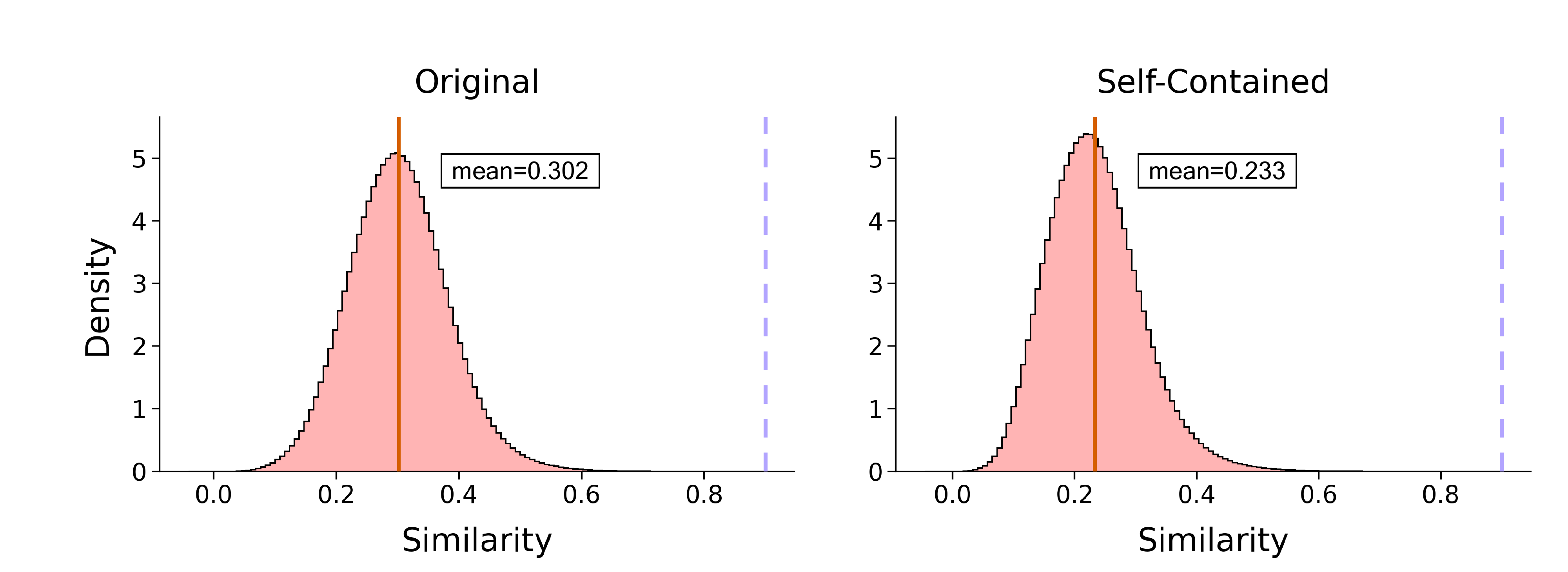}}{\fbox{\parbox{0.95\textwidth}{\centering\textbf{[Placeholder]} \texttt{figures/similarity\_distribution.pdf} not found.}}}
\captionsetup{hypcap=false}
\captionof{figure}{\footnotesize \textbf{Pairwise Similarity Distribution.} Empirical distributions of pairwise embedding similarities across all $98{,}778{,}540$ problem pairs. Dashed vertical lines indicate the maximum similarity values, both below the $0.90$ duplicate threshold. \emph{Left}: similarities between original statements. \emph{Right}: similarities between self-contained rewrites.}
\label{fig:similarity-distribution}
\end{center}
\vspace{0.75em}

\section{Near-Duplicates Filtering Details}
\label{app:filtering-details}

\subsection{Pairwise Similarity Distribution}
\label{app:sim-distribution}
\paragraph{}
Figure~\ref{fig:similarity-distribution} reports the pairwise embedding similarity distributions for the original statements and the self-contained rewrites on the original collection of 20,835 problems.

\subsection{\texttt{GPT-5.5} Judgments Near Decision Boundary}

\label{app:gpt5.5-judges}
\paragraph{}
Tables~\ref{tab:boundary-judgement-1}, \ref{tab:boundary-judgement-2}, and \ref{tab:boundary-judgement-3} present problem pairs with similarity scores close to the $0.9$ threshold that \texttt{GPT-5.5} nevertheless judged to be distinct. These cases support our conservative threshold choice.

\clearpage
\onecolumn

\begin{tcblisting}{
  enhanced,
  breakable,
  listing only,
  colback=teal!04,
  colframe=teal!25,
  coltitle=black,
  fonttitle=\small\bfseries\color{black},
  sharp corners,
  boxrule=0.6pt,
  left=1.2mm,right=1.2mm,top=1.0mm,bottom=1.0mm,
  before skip=6pt, after skip=2pt,
  title={\texttt{GPT-5.5} Judgment Example 1 ($similarity=0.89967543$)},
  listing options={
    basicstyle=\ttfamily\scriptsize,
    breaklines=true,
    breakatwhitespace=false,
    columns=fullflexible,
    keepspaces=true,
    showstringspaces=false,
    escapeinside={(*@}{@*)}
  }
}
(*@\textbf{\texttt{[Problem 1]}}@*)
Let S be a convex body in R^3 (a compact convex set with non-empty interior). A convex body K ⊂ R^3
is called S-Kakeya if, for every rotation ρ ∈ SO(3), K contains a congruent copy ρ(S) + v of S (for
some translation vector v depending on ρ).

Say that S has the Kakeya rotation property if, for every convex S-Kakeya body K ⊂ R^3 and every two
congruent copies S₁, S₂ of S contained in K, there exists a continuous path of congruent copies of
S, all contained in K, starting at S₁ and ending at S₂ (so that any two copies of S inside K can be
continuously rotated and translated into one another while remaining inside K).

Does there exist a convex body S in R^3 that fails the Kakeya rotation property? Equivalently, can
one find S ⊂ R^3 and a convex S-Kakeya body K ⊂ R^3 containing two congruent copies of S that cannot
be joined by a continuous family of congruent copies of S lying inside K?

(*@\textbf{\texttt{[Problem 2]}}@*)
Let d ≥ 2 be an integer. For a convex body S ⊂ ℝ^d (a compact convex set with non-empty interior,
or, more generally, including line segments), say that a convex body K ⊂ ℝ^d is S-Kakeya if, for
every rotation ρ ∈ SO(d), K contains a translate of ρ(S). Say that S has the Kakeya rotation
property if, for every S-Kakeya convex body K ⊂ ℝ^d and every two congruent copies S₁, S₂ of S
contained in K, there exists a continuous family (S_t)_{t ∈ [0,1]} of congruent copies of S, all
contained in K, with S_0 = S₁ and S_1 = S₂.

It is known that, in every dimension d ≥ 2, line segments have the Kakeya rotation property, and
that d-dimensional Euclidean balls trivially have this property.

Question: For some dimension d ≥ 3, does there exist a convex body S ⊂ ℝ^d, other than a line
segment or a d-dimensional ball, that has the Kakeya rotation property?

(*@\textbf{\texttt{[GPT-5.5]}}@*)
They are **different problems**.

Problem 1 asks whether there exists a convex body (S \subset \mathbb{R}^3) that **fails** the Kakeya rotation property.

Problem 2 asks whether, in some dimension (d \ge 3), there exists a nontrivial convex body (S) that **has** the Kakeya rotation property.

So they ask about opposite kinds of examples: failure vs existence of nontrivial success.

\end{tcblisting}
\begin{center}
\captionsetup{type=table,hypcap=false}
\caption{\texttt{GPT-5.5} Judgment Example 1}
\label{tab:boundary-judgement-1}
\end{center}

\begin{tcblisting}{
  enhanced,
  breakable,
  listing only,
  colback=teal!04,
  colframe=teal!25,
  coltitle=black,
  fonttitle=\small\bfseries\color{black},
  sharp corners,
  boxrule=0.6pt,
  left=1.2mm,right=1.2mm,top=1.0mm,bottom=1.0mm,
  before skip=6pt, after skip=2pt,
  title={\texttt{GPT-5.5} Judgment Example 2 ($similarity=0.8989585$)},
  listing options={
    basicstyle=\ttfamily\scriptsize,
    breaklines=true,
    breakatwhitespace=false,
    columns=fullflexible,
    keepspaces=true,
    showstringspaces=false,
    escapeinside={(*@}{@*)}
  }
}
(*@\textbf{\texttt{[Problem 1]}}@*)
Consider the quadratic polynomial family
$$P_\alpha(z) = e^{2\pi i \alpha} z + z^2, \qquad z \in \mathbb{C},$$
where $\alpha \in \mathbb{R}\setminus\mathbb{Q}$. Recall that an irrational number $\alpha$ with
continued fraction convergents $p_n/q_n$ is called a Brjuno (Bruno) number if
$$\sum_{n=0}^{\infty} \frac{\log q_{n+1}}{q_n} < \infty.$$
By a theorem of Brjuno–Yoccoz, for every Brjuno number $\alpha$ the polynomial $P_\alpha$ is locally
linearizable at the fixed point $0$; that is, there is a maximal open neighborhood $S_\alpha$ of $0$
(the Siegel disk of $P_\alpha$) on which $P_\alpha$ is conformally conjugate to the rigid rotation
$z \mapsto e^{2\pi i \alpha} z$.

Question: Does there exist a Brjuno number $\alpha$ such that the boundary $\partial S_\alpha$ of
the Siegel disk of $P_\alpha$ is not a Jordan curve (i.e., is not homeomorphic to a circle)?

(*@\textbf{\texttt{[Problem 2]}}@*)
For an irrational real number $\alpha$, consider the quadratic polynomial
$$P_\alpha(z) = e^{2\pi i \alpha} z + z^2.$$
The Bruno series of $\alpha$ is defined by $B(\alpha) = \sum_{n \ge 0} \frac{\log q_{n+1}}{q_n}$,
where $p_n/q_n$ are the convergents of the continued fraction expansion of $\alpha$; the number
$\alpha$ is called a Bruno number if $B(\alpha) < \infty$. By a theorem of Bruno and Yoccoz,
$\alpha$ is a Bruno number if and only if $P_\alpha$ is analytically linearizable near $0$, i.e.,
there is a holomorphic injection $\varphi$ from a neighborhood of $0$ in $\mathbb{C}$ to
$\mathbb{C}$ with $\varphi(0)=0$, $\varphi'(0)=1$, and $\varphi^{-1} \circ P_\alpha \circ \varphi(z)
= e^{2\pi i \alpha} z$. Let $S_\alpha$ denote the maximal such linearization domain (the Siegel disk
of $P_\alpha$ around $0$).

Question: Does there exist a Bruno number $\alpha$ such that the boundary $\partial S_\alpha \subset
\mathbb{C}$ is a $C^\infty$ embedded circle (i.e., a $C^\infty$ submanifold of $\mathbb{C}$
diffeomorphic to $S^1$)?

(*@\textbf{\texttt{[GPT-5.5]}}@*)
They are **different problems**.

Both use the same quadratic family and the same Bruno/Brjuno condition, but the questions ask different things:

* **Problem 1:** asks whether some Siegel disk boundary is **not a Jordan curve**.
* **Problem 2:** asks whether some Siegel disk boundary is a **(C^\infty) embedded circle**, which is much stronger than being Jordan.

So they are not the same; they ask essentially opposite types of boundary regularity.

\end{tcblisting}
\begin{center}
\captionsetup{type=table,hypcap=false}
\caption{\texttt{GPT-5.5} Judgment Example 2}
\label{tab:boundary-judgement-2}
\end{center}

\begin{tcblisting}{
  enhanced,
  breakable,
  listing only,
  colback=teal!04,
  colframe=teal!25,
  coltitle=black,
  fonttitle=\small\bfseries\color{black},
  sharp corners,
  boxrule=0.6pt,
  left=1.2mm,right=1.2mm,top=1.0mm,bottom=1.0mm,
  before skip=6pt, after skip=2pt,
  title={\texttt{GPT-5.5} Judgment Example 3 ($similarity=0.89961576$)},
  listing options={
    basicstyle=\ttfamily\scriptsize,
    breaklines=true,
    breakatwhitespace=false,
    columns=fullflexible,
    keepspaces=true,
    showstringspaces=false,
    escapeinside={(*@}{@*)}
  }
}
(*@\textbf{\texttt{[Problem 1]}}@*)
Let $D=\{z\in\mathbb{C}:|z|<1\}$ be the open unit disc, and let $n\ge 3$ be an integer. Consider the
family
$$\mathcal{F} = \{\, f : f \text{ is meromorphic on } D \text{ and } f'(z)\,f(z)^n \neq 1 \text{ for
   every } z\in D\,\}.$$
Prove or disprove: the family $\mathcal{F}$ is normal on $D$ (in the sense of Montel, with respect
to the spherical metric); that is, every sequence in $\mathcal{F}$ has a subsequence which converges
locally uniformly on $D$ to a meromorphic function or to the constant $\infty$.

(*@\textbf{\texttt{[Problem 2]}}@*)
Let $\mathbb{D}=\{z\in\mathbb{C}:|z|<1\}$ denote the open unit disc, and fix an integer $n\ge 1$.
Let $\mathcal{F}_n$ denote the family of all functions $f$ that are holomorphic in $\mathbb{D}$ and
satisfy
\[ f'(z)\,f(z)^{n}\neq 1 \qquad\text{for every } z\in\mathbb{D}. \]
A family of holomorphic functions on $\mathbb{D}$ is called normal if every sequence in the family
has a subsequence that converges locally uniformly on $\mathbb{D}$ either to a holomorphic function
or to $\infty$.

Prove or disprove: for every integer $n\ge 1$, the family $\mathcal{F}_n$ is normal in $\mathbb{D}$.

(*@\textbf{\texttt{[GPT-5.5]}}@*)
They are **different problems**.

Key differences:

1. **Function class:**
   Problem 1 allows **meromorphic** functions; Problem 2 only allows **holomorphic** functions.

2. **Range of (n):**
   Problem 1 assumes (n \ge 3); Problem 2 asks for **all (n \ge 1)**.

3. **Normality target:**
   Problem 1 allows subsequential limits to be **meromorphic or (\infty)** in the spherical metric.
   Problem 2 asks for holomorphic normality, with limits **holomorphic or (\infty)**.

So they are related, but **not the same problem**.
\end{tcblisting}
\begin{center}
\captionsetup{type=table,hypcap=false}
\caption{\texttt{GPT-5.5} Judgment Example 3}
\label{tab:boundary-judgement-3}
\end{center}

\clearpage

\section{Reasoning Behavior Details}
\label{app:behaviour-details}
\label{app:per-model-refs}

\subsection{Eight-Model Study on \dataset{}}
\label{app:eight-model-study}
\label{sec:behavioral-metrics}

This subsection reports the corpus-level behavioral study referenced in Section~\ref{sec:results}: eight open-weight models evaluated on \dataset{} and four control benchmarks with two complementary metric families.

\paragraph{Behavior and Factuality Metrics.}~We use two complementary methods. Rule-based counting matches three curated phrase lists for citation-like mentions (\texttt{cite}), abandonment (\texttt{abandon}), and claims made without justification (\texttt{assume}) against the lowercased trace. Per benchmark, we report the fraction of traces with at least one match. Agent-based reference verification uses a two-stage pipeline that extracts reference-like spans with \texttt{GPT-5.4-nano} and checks each span against web search using a \texttt{Codex} agent. The counters identify surface patterns, while reference verification checks whether cited sources exist. Neither method establishes mathematical correctness. Full keyword lists, verification prompts, and aggregation details are in Appendix~\ref{app:keyword-judge-metric}.

\paragraph{Setup.}~We evaluate a broad set of models, including several substantially larger systems and both older and newer generations from each model family: \texttt{DeepSeek R1}~\citep{deepseekr1}, \texttt{DeepSeek V4-Pro}~\citep{deepseekai2026deepseekv4}, \texttt{Kimi K2}~\citep{team2025kimi}, \texttt{Kimi K2.6}~\citep{team2026kimi}, \texttt{Qwen3-30B-A3B-Thinking} and \texttt{Qwen3-235B-A22B}~\citep{yang2025qwen3}, and \texttt{Qwen3.5} (35B-A3B, 397B-A17B)~\citep{qwen3.5}. Throughout the analysis we group these models into four older$\to$newer matched pairs (\texttt{R1}$\to$\texttt{V4-Pro}, \texttt{K2}$\to$\texttt{K2.6}, \texttt{Qwen3-30B-A3B-Thinking}$\to$\texttt{Qwen3.5 35B}, and \texttt{Qwen3 235B}$\to$\texttt{Qwen3.5 397B}). Each model answers $90$ problems sampled from \dataset{} and from each of the four control benchmarks ($86$ for SOOHAK; sampling details in Appendix~\ref{app:benchmark-details}), for $446$ prompts per model.

\begin{figure}[t]
\centering
\includegraphics[width=0.95\textwidth]{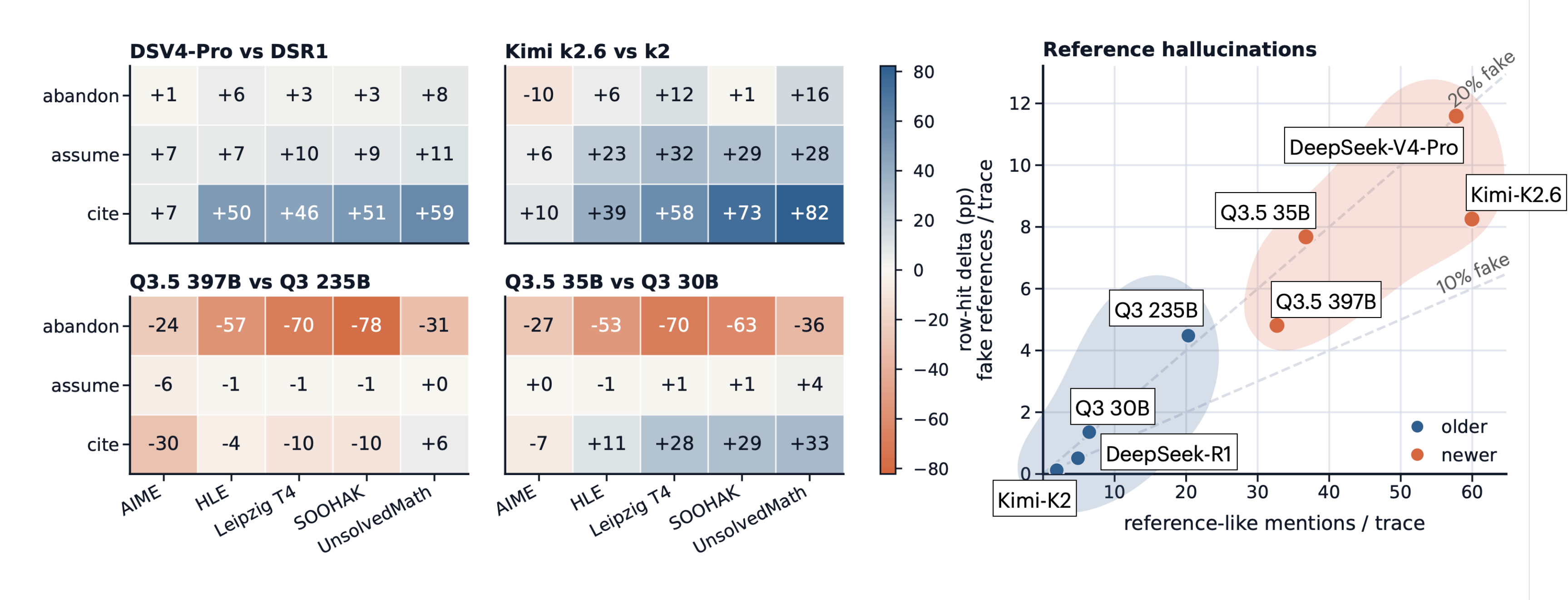}
\caption{\footnotesize
\textbf{Citation behavior across four matched older$\to$newer model pairs.} \textbf{Left}: Row-hit-rate deltas (newer minus older, in percentage points) for the three rule-based counters (\texttt{abandon}, \texttt{assume}, \texttt{cite}) across the five benchmarks. \textbf{Right}: Agent-Judge reference verification on 720 \dataset{} traces, one point per model. Newer models (orange) sit upper-right of their predecessors (blue), with more reference-like mentions per trace (x-axis) and more references judged fake per trace (y-axis); dashed guides mark 10\% and 20\% fake-reference shares. Full per-model counts and fake-reference rates in Table~\ref{tab:per-model-refs}.}
\label{fig:reference-refinement-line}
\end{figure}

\begin{tcolorbox}[
  breakable,
  enhanced,
  colback=teal!04,
  colframe=teal!25,
  coltitle=black,
  fonttitle=\small\bfseries\color{black},
  fontupper=\footnotesize,
  sharp corners,
  boxrule=0.6pt,
  left=1.2mm,right=1.2mm,top=1.0mm,bottom=1.0mm,
  before skip=6pt, after skip=6pt
]
\textbf{Finding}: Across model families, the newer generation cites external sources more often than the older one, and produces more fake references per trace.
\end{tcolorbox}

\paragraph{Citation Behavior.}~Citation-like reasoning rises sharply in newer model generations (Figure~\ref{fig:reference-refinement-line}, left, \texttt{cite} row), with row-hit rates increasing by 30-80 percentage points on \dataset{} across the DeepSeek, Kimi, and Qwen3 matched pairs. The same pattern holds on the other research-level benchmarks and weakens as benchmarks get easier (Appendix~\ref{app:cross-benchmark}), suggesting that newer models' tendency to cite is an artifact of the academic level of the questions.

To supplement the keyword counter, we use the Agent-Judge pipeline described above on 90 traces from \dataset{} for each of our 8 models. Across all 720 traces, 629 (87.4\%) cite at least one reference-like object and 389 (54.0\%) contain at least one fake reference. At the reference level, we inspect 19,864 extracted mentions and label 3,492 fake (17.6\%) after consulting internet search (Figure~\ref{fig:reference-refinement-line}, right). Per-trace mention counts grow dramatically across the matched comparisons. DeepSeek R1 $\to$ V4-Pro rises from $4.9$ to $57.8$ mentions per trace ($0.5\to11.6$ fakes), Kimi K2 $\to$ K2.6 from $1.9$ to $60.0$ ($0.1\to8.3$ fakes), Qwen3-30B-A3B-Thinking $\to$ Qwen3.5 35B from $6.5$ to $36.7$ ($1.4\to7.7$ fakes), and Qwen3 235B $\to$ Qwen3.5 397B from $20.3$ to $32.7$ ($4.5\to4.8$ fakes). In aggregate, newer models produce $5.6\times$ more reference-like mentions per trace and $5.0\times$ more fakes. The fake mentions are mostly hallucinated paper titles and author attributions. Models try to ground their arguments on wrong statements by fabricating that a supporting reference exists, making the result sound correct. Representative fakes:
\begin{itemize}[leftmargin=1.2em, itemsep=2pt, topsep=4pt]
\setlength{\itemsep}{0pt}
\item ``Neeman's paper: \emph{A remark on the unique factorization theorem}''
\item ``J. Winkelmann, \emph{On the holomorphic equivalence of the Koras--Russell cubic}''
\item ``a specific paper: \emph{On the probability that a random polynomial is stable} by J. M. Anderson''
\end{itemize}

\begin{tcolorbox}[
  breakable,
  enhanced,
  colback=teal!04,
  colframe=teal!25,
  coltitle=black,
  fonttitle=\small\bfseries\color{black},
  fontupper=\footnotesize,
  sharp corners,
  boxrule=0.6pt,
  left=1.2mm,right=1.2mm,top=1.0mm,bottom=1.0mm,
  before skip=6pt, after skip=6pt
]
\textbf{Finding.} On \dataset{}, the \texttt{assume} counter matches $94.0\%$ of traces, and $54.0\%$ contain at least one reference judged fake.
\end{tcolorbox}

\paragraph{Interpreting Behavioral Signals.}~It should be noted, however, that citations and compression are not themselves failures. Mathematicians cite, reduce, and skip routine details too, and if models could ground their citations correctly, this would be less of a concern. But citations are not the only place where models try to look the part. On \dataset{} the \texttt{abandon} counter matches only $125/720$ traces ($17.4\%$), while \texttt{assume} matches $677/720$ ($94.0\%$; Table~\ref{tab:behavior-counter-rates}). Models rarely give up outright, and the attempt almost always leans on compressed claims rather than from-scratch derivation. These surface signs resemble mathematician practice, but we cannot tell whether models employ the underlying reasoning or simply parrot the form.

\subsection{Our Conjecture: Why Do Newer Models Fabricate More?}
\label{app:conjecture}
\label{sec:conjecture}

Interestingly, we observe that models released in 2025 (\texttt{DeepSeek R1}, \texttt{Kimi K2}, \texttt{Qwen3}) cite less, while models released in 2026 (\texttt{DeepSeek V4-Pro}, \texttt{Kimi K2.6}, \texttt{Qwen3.5}) cite far more, with more fake citations. The per-pair \emph{rates} are more nuanced: the fake share among reference-like mentions rises for DeepSeek ($10.2\%\to20.1\%$) and Kimi ($5.3\%\to13.8\%$), stays roughly flat for \texttt{Qwen3-30B-A3B-Thinking}$\to$\texttt{Qwen3.5 35B} ($21.5\%\to21.0\%$), and decreases for \texttt{Qwen3 235B}$\to$\texttt{Qwen3.5 397B} ($22.2\%\to14.7\%$). What increases consistently is \emph{exposure}: because newer models cite so much more, the number of fabricated references a user encounters per response grows in every pair, several-fold in three of the four, even where the conditional fake rate does not (per-model values in Table~\ref{tab:per-model-refs}). Because this citation-density shift holds across DeepSeek, Kimi, and Qwen, three different model families, it is unlikely to be a quirk of any single training set. One plausible explanation is internet-search RL, or more broadly agentic RL~\citep{dong2025agentic,liu2025webexplorer,li2026literesearcher}. Recent post-training pipelines often place the model inside an agentic harness at train time, equipped with explicit search and citation tools, and reward it for grounding claims in retrieved sources. Over training, the model learns to invoke papers, books, and URLs as a routine part of producing an authoritative-looking answer. In our setting, however, models are evaluated without internet access. A plausible explanation is that rather than abandoning the citation behavior when the search tool is unavailable, models keep invoking the learned pattern and simply fabricate the references they would normally retrieve. We note that no intervention in our study isolates this factor from other differences between model generations, such as scale, data, or post-training recipe, so we present it as a hypothesis rather than a conclusion.

\subsection{Behavior-Counter Rates}
\label{app:behavior-counter-rates}

Table~\ref{tab:behavior-counter-rates} reports the fraction of traces in each benchmark that trigger each behavior counter across the eight evaluated models.

\begin{center}
\begin{minipage}{\textwidth}
\centering
\footnotesize
\setlength{\tabcolsep}{2pt}
\captionsetup{hypcap=false}
\captionof{table}{\footnotesize Absolute behavior-counter row-hit rates across the eight paper models. The \texttt{assume}, \texttt{cite}, and \texttt{abandon} counters correspond to the three rule-based keyword groups defined in Appendix~\ref{app:keyword-groups}.}
\label{tab:behavior-counter-rates}
\resizebox{\textwidth}{!}{%
\begin{tabular}{lccc}
\toprule
Benchmark & \texttt{assume} rows & \texttt{cite} rows & \texttt{abandon} rows \\
\midrule
\dataset{} & $677/720$ ($94.0\%$) & $482/720$ ($66.9\%$) & $125/720$ ($17.4\%$) \\
Leipzig Tier-4 & $666/720$ ($92.5\%$) & $461/720$ ($64.0\%$) & $200/720$ ($27.8\%$) \\
SOOHAK & $651/688$ ($94.6\%$) & $443/688$ ($64.4\%$) & $161/688$ ($23.4\%$) \\
HLE-Verified & $673/720$ ($93.5\%$) & $272/720$ ($37.8\%$) & $141/720$ ($19.6\%$) \\
AIME & $660/720$ ($91.7\%$) & $72/720$ ($10.0\%$) & $70/720$ ($9.7\%$) \\
\midrule
Research-level total & $1{,}994/2{,}128$ ($93.7\%$) & $1{,}386/2{,}128$ ($65.1\%$) & $486/2{,}128$ ($22.8\%$) \\
\bottomrule
\end{tabular}
}%
\end{minipage}
\end{center}
\vspace{0.75em}

\subsection{Cross-Benchmark Patterns}
\label{app:cross-benchmark}

The patterns reported on \dataset{} in Appendix~\ref{app:eight-model-study} replicate across the control benchmarks. The \texttt{cite} row-hit rate rises by $30$--$80$ percentage points on Leipzig Tier-4 and SOOHAK across the DeepSeek, Kimi, and Qwen3 matched pairs, weakens on HLE, and is near zero on AIME (Figure~\ref{fig:reference-refinement-line}, left), consistent with citation behavior being tied to the academic level of the questions. Aggregate behavior-counter rates per benchmark are given in Table~\ref{tab:behavior-counter-rates}.

\subsection{Per-Model Reference Verification}
\label{app:per-model-reference-verification}

Table~\ref{tab:per-model-refs} reports the Agent-Judge reference-verification statistics per model on the $90$ \dataset{} traces per model from Appendix~\ref{app:eight-model-study}: reference-like mentions per trace, references judged fake per trace, and the resulting fake share. The fake share rises for the DeepSeek and Kimi pairs, stays roughly flat for the smaller Qwen pair, and decreases for the larger Qwen pair, while mentions per trace rise for all four pairs.

\begin{center}
\begin{minipage}{\textwidth}
\centering
\footnotesize
\setlength{\tabcolsep}{4pt}
\captionsetup{hypcap=false}
\captionof{table}{\footnotesize \textbf{Per-model reference verification on \dataset{}.} Reference-like mentions and judged-fake references per trace over $90$ traces per model, with the fake share (fakes over mentions).}
\label{tab:per-model-refs}
\begin{tabular}{lccc}
\toprule
Model & Mentions / trace & Fakes / trace & Fake share \\
\midrule
\texttt{DeepSeek R1} & $4.9$ & $0.5$ & $10.2\%$ \\
\texttt{DeepSeek V4-Pro} & $57.8$ & $11.6$ & $20.1\%$ \\
\texttt{Kimi K2} & $1.9$ & $0.1$ & $5.3\%$ \\
\texttt{Kimi K2.6} & $60.0$ & $8.3$ & $13.8\%$ \\
\texttt{Qwen3-30B-A3B-Thinking} & $6.5$ & $1.4$ & $21.5\%$ \\
\texttt{Qwen3.5 35B-A3B} & $36.7$ & $7.7$ & $21.0\%$ \\
\texttt{Qwen3 235B-A22B} & $20.3$ & $4.5$ & $22.2\%$ \\
\texttt{Qwen3.5 397B-A17B} & $32.7$ & $4.8$ & $14.7\%$ \\
\bottomrule
\end{tabular}
\end{minipage}
\end{center}

\section{License and Release}
\label{app:license}

We release the \textsc{ResearchMath} family under the MIT License. The release includes \dataset{}, the corpus of $14{,}056$ research-level mathematical problems described in Section~\ref{sec:collection}, the full $220$K \datareasoning{} collection described in Section~\ref{sec:generating-response}, and the $77$K training subset constructed through length filtering, keyword filtering, and random sampling as detailed in Appendix~\ref{app:keyword-groups}. These artifacts derive from publicly available academic sources (arXiv preprints, open-problem web pages, and workshop or conference problem sheets). The Extractor agent discards any document hidden behind a paywall or a non-permissive license before extraction (Section~\ref{sec:collection-process}); paywalled or restricted sources are not represented in the released data.

\section{Training Details}
\label{app:training}

\paragraph{Full-parameter SFT.}~The main student checkpoints are \texttt{SmolLM3-3B IT-LC-Expert}, \texttt{Qwen3-4B-Instruct-2507}, and \texttt{Olmo-3-7B-Instruct}, as described in Section~\ref{sec:learning-setup}. Table~\ref{tab:main-results-30b} additionally reports results for \texttt{Qwen3-30B-A3B-Instruct-2507}. All full-parameter runs use the training configuration described below.

\paragraph{Optimization and batching.}~We use fused PyTorch AdamW with a peak learning rate of $5\times10^{-5}$, $(\beta_1,\beta_2)=(0.9,0.999)$, $\epsilon=10^{-8}$, no weight decay, and a maximum gradient norm of $1.0$. Training uses BF16 mixed precision for six epochs, with a cosine learning-rate schedule, $3\%$ warmup, and a minimum learning rate of $10^{-5}$. The effective global batch size is $64$ sequences across eight GPUs. Each device normally processes one sequence per microbatch, with eight gradient-accumulation steps. The Qwen3-4B Nemotron run uses two sequences per device and four accumulation steps, preserving the same global batch size.

\paragraph{Sequence preparation and objective.}~We concatenate tokenized examples offline into fixed blocks of $65{,}536$ tokens and discard the final incomplete block. The trainer's own packing and dataset-preparation stages are disabled because the data are already packed. No additional truncation is applied at training time. We apply the causal language-model objective to the packed sequences, including problem text, reasoning, final answers, and chat-template tokens. Prompt tokens are not masked from the loss, and attention is not isolated across example boundaries within a block. Loss normalization uses token counts aggregated across devices.

\paragraph{Seeds and checkpoint selection.}~Each of the ten recorded configurations is trained once with seed $42$, which is also used for shuffling during dataset construction. These records contain no independent training-seed replicates. We evaluate the final checkpoint after the sixth epoch.

\paragraph{Implementation.}~We use FSDP full sharding across eight GPUs on a single node, with transformer-layer wrapping and no parameter offloading. We use a mix of H100 and H200 GPUs across training runs. We enable FlexAttention, Liger kernels, and reentrant gradient checkpointing through Hugging Face, with FSDP activation checkpointing disabled. FSDP's \texttt{use\_orig\_params} setting is disabled for the $30$B runs and enabled for the smaller models. Table~\ref{tab:full-sft-software} lists the software versions recorded for the two environments.

\begin{center}
\small
\setlength{\tabcolsep}{8pt}
\captionsetup{hypcap=false}
\captionof{table}{\footnotesize \textbf{Software versions for full-parameter training.} Versions are taken from the run-time environment records.}
\label{tab:full-sft-software}
\begin{tabular}{@{}lcc@{}}
\toprule
Package & $30$B runs & $3$B/$4$B/$7$B runs \\
\midrule
PyTorch & \texttt{2.10.0+cu128} & \texttt{2.11.0} \\
Transformers & \texttt{4.57.6} & \texttt{5.13.0} \\
TRL & \texttt{0.11.4} & \texttt{1.9.2} \\
Accelerate & \texttt{1.13.0} & \texttt{1.12.0} \\
Liger Kernel & \texttt{0.8.0} & \texttt{0.8.1} \\
\bottomrule
\end{tabular}
\end{center}

\begin{center}
\scriptsize
\setlength{\tabcolsep}{5pt}
\captionsetup{hypcap=false}
\captionof{table}{\footnotesize \textbf{Full-parameter SFT results for Qwen3-30B-A3B-Instruct-2507.} We report mean accuracy and standard deviation over three runs in a stochastic setting. Bold denotes the best result for each benchmark.}
\label{tab:main-results-30b}
\begin{tabular}{@{}lcccc@{}}
\toprule
\multicolumn{1}{c}{\raisebox{-1.7ex}[0pt][0pt]{Model}} & \multicolumn{1}{c}{Olympiad Level} & \multicolumn{3}{c}{Graduate/Research} \\
\cmidrule(lr){2-2}\cmidrule(lr){3-5}
 & AIME 24-26 & HLE & SOOHAK & ArxivMath \\
\midrule
Qwen3-30B-A3B-Instruct-2507 & $71.85{\scriptstyle\,\pm\,1.7}$ & $13.97{\scriptstyle\,\pm\,1.8}$ & $45.91{\scriptstyle\,\pm\,1.4}$ & TBD \\
\quad + Nemotron-SFT-Math-v4 & $78.15{\scriptstyle\,\pm\,2.8}$ & $\mathbf{20.32}{\scriptstyle\,\mathbf{\pm\,1.1}}$ & $44.58{\scriptstyle\,\pm\,1.3}$ & TBD \\
\quad + DASD & $\mathbf{80.74}{\scriptstyle\,\mathbf{\pm\,1.3}}$ & $18.31{\scriptstyle\,\pm\,0.2}$ & $46.71{\scriptstyle\,\pm\,0.2}$ & TBD \\
\quad + ResearchMath (Ours) & $75.19{\scriptstyle\,\pm\,1.3}$ & $18.62{\scriptstyle\,\pm\,1.0}$ & $\mathbf{48.30}{\scriptstyle\,\mathbf{\pm\,0.7}}$ & TBD \\
\bottomrule
\end{tabular}
\end{center}

\clearpage

\subsection[LoRA Training on 5K Trajectories]{LoRA Training on $5$K Trajectories}
\label{app:lora-results}

We also test the mitigated trajectories from Section~\ref{sec:mitigation} with smaller training sets and LoRA~\citep{hu2021lora}. We fine-tune three Qwen3 base models (\texttt{Qwen3-4B-base}, \texttt{Qwen3-8B-base}, and \texttt{Qwen3-30B-A3B-base}) on $5{,}000$ randomly sampled traces from either the mitigated \datareasoning{} generations or the DASD-Thinking control.\footnote{LoRA targets all linear layers in the \texttt{4B} and \texttt{8B} models, but we targeted only attention projections in \texttt{30B-A3B} by accident. We retain these runs because they still improve over the base model on all three benchmarks.} Each setting uses three seeds, and reported scores are means over the runs. Table~\ref{tab:finetuning-eval-results} reports the complete scores and standard deviations, including the matched DASD controls.

Training on mitigated \textsc{ResearchMath} trajectories yields a $9.2$-point gain over the base models, averaged across models and benchmarks. An early run fine-tuning \texttt{Qwen3-4B} on unfiltered \texttt{Qwen3-30B-A3B-Thinking} responses scored close to zero on nearly every benchmark. We omit this failed run from Table~\ref{tab:finetuning-eval-results} because it produced degenerate outputs. These results support prompted and filtered research-level trajectories as post-training data.

\begin{table}[ht]
\centering
\scriptsize
\setlength{\tabcolsep}{4pt}
\caption{\footnotesize \textbf{LoRA fine-tuning results across model sizes and benchmarks.} Each fine-tuned variant uses $5{,}000$ trajectories from DASD or \textsc{ResearchMath}. Scores are means with standard deviations across three seeds, in percentage points. \textsc{ResearchMath} cells are shaded green where they exceed the DASD control and red where they fall below it.}
\label{tab:finetuning-eval-results}
\begin{tabular}{@{}lccc@{}}
\toprule
System & AIME & HLE & SOOHAK \\
\midrule
\texttt{Qwen3-4B-base} & $1.1_{\pm1.1}$ & $4.3_{\pm2.5}$ & $8.3_{\pm5.0}$ \\
\quad + DASD & $10.0_{\pm1.1}$ & $4.9_{\pm1.0}$ & $20.8_{\pm0.3}$ \\
\quad + \dataset{} & \cellcolor{green!15}$11.9_{\pm2.6}$ & \cellcolor{green!15}$7.0_{\pm1.1}$ & \cellcolor{green!15}$24.6_{\pm0.7}$ \\
\midrule
\texttt{Qwen3-8B-base} & $1.1_{\pm0.6}$ & $4.8_{\pm2.9}$ & $7.4_{\pm4.3}$ \\
\quad + DASD & $7.8_{\pm1.1}$ & $4.4_{\pm0.3}$ & $19.6_{\pm2.0}$ \\
\quad + \dataset{} & \cellcolor{green!15}$13.3_{\pm2.8}$ & \cellcolor{green!15}$5.6_{\pm0.8}$ & \cellcolor{green!15}$22.2_{\pm0.3}$ \\
\midrule
\texttt{Qwen3-30B-A3B-base} & $3.0_{\pm1.3}$ & $4.0_{\pm0.2}$ & $14.2_{\pm0.3}$ \\
\quad + DASD & $24.4_{\pm14.2}$ & $3.5_{\pm2.0}$ & $23.2_{\pm1.3}$ \\
\quad + \dataset{} & \cellcolor{red!15}$13.3_{\pm8.0}$ & \cellcolor{green!15}$7.9_{\pm4.6}$ & \cellcolor{green!15}$25.0_{\pm1.4}$ \\
\bottomrule
\end{tabular}
\end{table}

\paragraph{LoRA configuration.} We use rank $r=64$, alpha $\alpha=128$, dropout $0.05$, and no bias. For the \texttt{4B} and \texttt{8B} models, LoRA targets all linear layers in each transformer block, including attention projections (\texttt{q\_proj}, \texttt{k\_proj}, \texttt{v\_proj}, \texttt{o\_proj}) and MLP projections (\texttt{gate\_proj}, \texttt{up\_proj}, \texttt{down\_proj}). For the \texttt{30B-A3B} runs, we targeted only the attention projections by accident. 

\paragraph{Batching.} Per-device batch size is $1$; global batch size is $16$ for the $30$B run and $32$ for the smaller models. 

\paragraph{Sequence length.} Examples are truncated at $24{,}512$ tokens for the $4$B/$8$B runs and $32{,}768$ tokens for the $30$B run.

\clearpage

\section[Trajectory Correctness and Diversity]{Trajectory Correctness and Diversity}
\label{app:correctness-diversity}

\paragraph{Samples and judgments.}~For correctness assessment, we start from random samples of $10{,}000$ trajectories from each of \textsc{ResearchMath} $77$K, DASD $50$K, and \texttt{Nemotron-SFT-Math-v4} $86$K. Both the correctness and diversity analyses use \textsc{ResearchMath} trajectories drawn from the final $77$K training set constructed in Appendix~\ref{app:keyword-groups}. Processing completes for $9{,}968$ Nemotron trajectories, excluding the $32$ longest. From the processed samples, we draw $1{,}000$ trajectories per dataset with seed $2026$ for correctness assessment. Each judge receives the problem and raw reasoning trace at temperature $0$ and assigns one of four labels to the approach the trace commits to. Abandoned false starts and backtracking are explicitly excluded as defects. Table~\ref{tab:trajectory-correctness} reports the counts. Rates in Section~\ref{sec:trajectory-correctness} and Table~\ref{tab:trajectory-judge-summary} exclude missing or unparseable verdicts. The judgments lack reference solutions for the research-level problems, and differences in difficulty limit comparisons across corpora.

Of the $1{,}000$ sampled \textsc{ResearchMath} trajectories, $994$ receive valid verdicts from both judges. Both label $628$ incorrect and $25$ correct, corresponding to $63.2\%$ and $2.5\%$ of jointly assessed trajectories. The six remaining samples lack a Claude verdict. Research-level difficulty may challenge both judges, and shared errors remain possible even when they agree. Appendix~\ref{app:approach-correctness-prompt} reproduces the prompt.

\begin{table}[ht]
\centering
\caption{Judged approach correctness for $1{,}000$ sampled trajectories per dataset and judge. Columns give counts of valid verdicts and each label. A correct label concerns the assessed approach and does not certify a solution to the underlying problem.}
\label{tab:trajectory-correctness}
{\small
\setlength{\tabcolsep}{4pt}
\begin{tabular}{@{}llrrrrr@{}}
\toprule
Dataset & Judge & Valid & Correct & Partially & Incorrect & Unverifiable \\
\midrule
\texttt{Nemotron-SFT-Math-v4} & Claude-Fable-5.1 & $1{,}000$ & $923$ & $57$ & $13$ & $7$ \\
 & GPT-5.6-Sol & $999$ & $777$ & $164$ & $57$ & $1$ \\
DASD & Claude-Fable-5.1 & $987$ & $815$ & $121$ & $48$ & $3$ \\
 & GPT-5.6-Sol & $999$ & $724$ & $143$ & $132$ & $0$ \\
\textsc{ResearchMath} & Claude-Fable-5.1 & $994$ & $43$ & $185$ & $651$ & $115$ \\
 & GPT-5.6-Sol & $1{,}000$ & $37$ & $71$ & $842$ & $50$ \\
\bottomrule
\end{tabular}}
\end{table}

\paragraph{Matched diversity comparison.}~For diversity analysis, we sample $4{,}000$ trajectories from each dataset without applying a math-only filter to DASD. We then form $3{,}893$ triplets containing one trajectory per dataset, matched on thought-unit counts from \textsc{ThinkProbe}~\citep{kerkouri2026thinkprobe}. Median thought-unit counts are $101$, $100$, and $101$ for Nemotron, DASD, and \textsc{ResearchMath}, respectively. Within-triplet spread is at most $10\%$ in $99.0\%$ of triplets. This controls thought-unit count rather than token length. The correctness sample is drawn separately and does not use thought-unit matching.

\paragraph{Embedding trajectories and questions.}~We use \texttt{Qwen3-Embedding-8B} through vLLM in embedding mode, with a maximum context length of $32{,}768$, one process per GPU, and batches of $1{,}500$. We apply the same procedure separately to raw reasoning and question text, without instruction prefixes, and align the resulting vectors by example identifier. Text is tokenized without special tokens. Texts of at most $30{,}000$ tokens are embedded whole, while longer texts are split into consecutive $30{,}000$-token chunks and decoded back to text for embedding. If chunk $j$ has embedding $e_j$ and token count $t_j$, the trajectory vector is
\[
v = \frac{\sum_j t_j e_j}{\left\|\sum_j t_j e_j\right\|_2}.
\]
This yields one $4{,}096$-dimensional unit vector per text, stored in \texttt{fp16}. Chunk averaging compresses within-trajectory variation, so the representation need not preserve the diversity of approaches within a long trace. Matching thought-unit counts does not eliminate differences in token length or chunk count.

\paragraph{Vendi score.}~Let $X\in\R^{n\times4{,}096}$ contain the unit vectors of a subsample. Its cosine similarity kernel is $K=XX^\top$. Following \citet{friedman2022vendi}, we compute
\[
V(X)=\exp\!\left(-\sum_i\lambda_i\log\lambda_i\right),
\qquad \lambda_i=\operatorname{eig}_i(K/n).
\]
The eigenvalues sum to $1$ in exact arithmetic. The score equals $1$ for identical vectors and $n$ for mutually orthogonal vectors, and can be interpreted as an effective number of distinct items under the chosen similarity measure. A score of $139.4$ therefore does not mean that exactly $139$ mathematical strategies were identified. The implementation computes eigenvalues of $X^\top X/n$, which shares all nonzero eigenvalues with $XX^\top/n$, avoiding construction of the full pairwise kernel. Eigenvalues at most $10^{-12}$ are discarded and the retained values are renormalized to sum to $1$. This spectrum is uncentered and describes similarity rather than covariance.

\paragraph{Effective dimensionality.}~We also compute the participation ratio of the centered sample covariance matrix $C$, retaining its positive eigenvalues $\mu_i$,
\[
D_{\mathrm{eff}}(X)=\frac{(\sum_i\mu_i)^2}{\sum_i\mu_i^2}.
\]
For nonzero covariance, this lies between $1$ and $\min(4{,}096,n-1)$ and summarizes the effective number of embedding directions contributing to variation. Unlike Vendi, it uses mean-centered vectors and a second-moment ratio rather than entropy, giving less weight to small eigenvalues. Agreement between the metrics indicates that the ranking holds under these two summaries, but both depend on the same embeddings and do not constitute independent evidence of mathematical strategy diversity.

\paragraph{Reasoning-to-question ratio and subsampling.}~For each of ten random $2{,}000$-trajectory subsamples per matched dataset, we compute $R/Q=V(X_{\mathrm{reasoning}})/V(X_{\mathrm{question}})$ using the same example identifiers in both matrices. Table~\ref{tab:trajectory-diversity-correctness} reports means and sample standard deviations over subsamples. Mean question Vendi scores are $150.70$, $165.47$, and $132.57$ for Nemotron, DASD, and \textsc{ResearchMath}, respectively. An R/Q ratio above $1$ means that reasoning embeddings have higher Vendi diversity than their corresponding question embeddings. Equal sample sizes make this a matched comparison, but do not guarantee that sample-size dependence cancels or that ratios are comparable across differently sized studies. The ratio does not establish that individual trajectories introduce novel mathematical ideas. Text length, style, and embedding geometry may affect it. Standard deviations describe subsampling variation within the fixed matched sets, not corpus-level uncertainty. These metrics measure embedding diversity rather than mathematical validity. Since the correctness sample is separate, the analysis does not isolate the diversity of incorrect trajectories or exclude a contribution from correct trajectories to downstream gains. Source, teacher, and construction differences also remain possible explanations.

\clearpage

\section{Prompts}
\label{app:prompts}

\subsection{Dataset Generation Agents}
\label{app:dataset-generation-agents}
Table~\ref{tab:extractor_prompt} and Table~\ref{tab:refiner_prompt} present the prompts used for the Extractor and Refiner agents in Section~\ref{sec:collection-process}, respectively.

\subsection{Difficulty Comparison}
\label{app:difficulty-comparision}
Table~\ref{tab:difficulty-comparision-prompt} shows the prompt for difficulty comparison in Section~\ref{sec:dataset-statistics}.

\subsection{Response Generation}
\label{app:response-generation}
Table~\ref{tab:response-prompt} shows the prompt used to generate model responses in Section~\ref{sec:generating-response}.
Table~\ref{tab:committed-attempt-prompt} presents the committed-attempt system prompt used for the prompt-level intervention and the released corpus.

\subsection{Factuality Metrics}
\label{app:factuality-metrics}
Tables~\ref{tab:fact-first-stage-prompt},~\ref{tab:fact-second-stage-prompt}, and~\ref{tab:fact-second-stage-short-block-prompt} present the prompts used for the factuality metric in Appendix~\ref{app:eight-model-study}. Table~\ref{tab:fact-second-stage-short-block-prompt} gives the short-block variant of Table~\ref{tab:fact-second-stage-prompt}, which is used for detected blocks shorter than 200 characters with additional surrounding reasoning context.

\subsection[Approach-Correctness Judging]{Approach-Correctness Judging}
\label{app:approach-correctness-prompt}
Both judges use the following prompt for the trajectory assessment in Section~\ref{sec:trajectory-correctness}.

\begin{tcblisting}{
  enhanced,
  breakable,
  listing only,
  colback=teal!04,
  colframe=teal!25,
  sharp corners,
  boxrule=0.6pt,
  left=1.2mm,right=1.2mm,top=1.0mm,bottom=1.0mm,
  before skip=6pt, after skip=2pt,
  listing options={
    basicstyle=\ttfamily\footnotesize\color{black},
    breaklines=true,
    breakatwhitespace=false,
    columns=fullflexible,
    keepspaces=true,
    showstringspaces=false
  }
}
You are an expert evaluator of mathematical and scientific reasoning traces.

You will be shown a PROBLEM and a model's raw REASONING TRACE (its private
working, not a polished answer). Traces are exploratory: false starts,
backtracking, and abandoned branches are normal and are NOT by themselves defects.

Judge APPROACH CORRECTNESS: is the approach the trace actually commits to sound?
Judge the reasoning, not merely the final number. An unfinished trace on the right
track is better than a completed one built on a false premise.

  correct       - the committed approach is valid and would reach a correct answer
  partially     - largely sound, but with a real gap, unjustified leap, or error
                  that does not invalidate the whole strategy
  incorrect     - the committed approach is invalid, circular, or rests on a false premise
  unverifiable  - an open/research problem with no checkable answer, or the trace
                  commits to nothing checkable

Reply with ONLY a JSON object, no prose, no code fence:
{"correctness":"correct|partially|incorrect|unverifiable","reason":"<=30 words"}
\end{tcblisting}

\clearpage
\onecolumn

\begin{tcblisting}{
  enhanced,
  breakable,
  listing only,
  colback=teal!04,
  colframe=teal!25,
  coltitle=black,
  fonttitle=\small\bfseries\color{black},
  sharp corners,
  boxrule=0.6pt,
  left=1.2mm,right=1.2mm,top=1.0mm,bottom=1.0mm,
  before skip=6pt, after skip=2pt,
  title={Prompt for the Extractor Agent},
  listing options={
    basicstyle=\ttfamily\scriptsize,
    breaklines=true,
    breakatwhitespace=false,
    columns=fullflexible,
    keepspaces=true,
    showstringspaces=false
  }
}
---
name: paper-question-parser
description: "Parse one paper or scholarly open-problem source (local PDF path, arXiv URL, direct PDF URL, DOI URL, HAL/DBLP/zbMATH/OpenAlex/Crossref landing page, workshop page, MathOverflow thread, or domain-specific Q&A forum), resolve it to readable full text, extract open/unsolved questions, rewrite them into self-contained form with evidence, dedupe, quality-gate, and auto-save structured JSON. Triggers: 'parse paper questions', 'extract open problems', 'unsolved questions from paper', 'parse open-problem page'."
---

# Paper Question Parser

You run a single-source workflow to extract open/unsolved research questions from a paper, workshop problem sheet, bibliographic landing page that resolves to full text, or scholarly problem page, and rewrite each into a self-contained, evidence-grounded question.

## Path Configuration
- **RUN_DIR**: `outputs/parse-paper/`
- **LATEST_FILE**: `outputs/latest.json`
- **TEMP_DIR**: `.sisyphus/paper-question-parser/tmp/`

## Operating Contract

- Process exactly one source per run.
- Accept either a local PDF path or one URL pointing to a scholarly source. Supported URL families include arXiv `abs`/`pdf`, direct PDFs, DOI URLs, HAL pages, DBLP/Crossref/OpenAlex/zbMATH landing pages, workshop/problem-session pages, and MathOverflow threads.
- Resolve URL inputs to a readable full-text artifact before extraction.
- Do not ask the user to run scripts.
- Return STRICT JSON only (no prose) with keys: `source`, `accepted`, `needs_review`, `trace`.
- Auto-save the final JSON result to disk every run.

## Input Normalization

1. Read `$ARGUMENTS` and extract a single source.
2. If input is a local PDF path:
   - Verify the file exists.
   - Use that absolute path as `paper_path`.
   - Set:
     - `source_kind = "pdf"`
     - `source_locator = paper_path`
     - `source_name = <local_pdf_stem>`
     - `citation_mode = "page_number"`
3. If input is a URL:
   - Normalize obvious wrappers while preserving the identity of the same source.
   - Resolve the URL in this order:
     1. If the URL is `https://arxiv.org/abs/<id>`, convert it to `https://arxiv.org/pdf/<id>.pdf`.
     2. If the URL is already a direct PDF URL, or the resolved response is a PDF, download it internally with Bash:
        - `mkdir -p TEMP_DIR`
        - `curl -L --fail "<pdf_url>" -o "TEMP_DIR/<paper_slug>.pdf"`
        - set `paper_path` to the downloaded file
        - set:
          - `source_kind = "pdf"`
          - `source_locator = paper_path`
          - `source_name = <arxiv_id_or_pdf_slug>`
          - `citation_mode = "page_number"`
     3. If the URL is a DOI URL or a bibliographic/landing page (for example Crossref, OpenAlex, DBLP, zbMATH, HAL, or a publisher landing page), inspect the page and follow the most specific full-text link available:
        - preferred targets: `PDF`, `Download article`, `Download extract`, `document`, `unpaywalled`, `external edition`, or an explicit article page whose visible body contains the problems
        - treat DBLP, Crossref, OpenAlex, and zbMATH primarily as resolvers, not as the technical source
        - for HAL, prefer an explicit `/document` or file-download link when available
        - if the landing page or download endpoint returns a missing-resource page, paywall page, or metadata-only page, query source-specific metadata APIs when available to recover the exact title and candidate readable copies of that same work:
          - HAL: `https://api.archives-ouvertes.fr/search/?q=halId_s:<id>&rows=1&fl=title_s,uri_s,openAccess_bool,submitType_s,label_xml&wt=json`
          - Crossref: `https://api.crossref.org/works/<doi>`
          - OpenAlex: `https://api.openalex.org/works?filter=doi:https://doi.org/<doi>&select=display_name,doi,open_access,best_oa_location,primary_location`
        - use API metadata only to identify the exact work, discover candidate full-text URLs, and detect access restrictions; do not extract open problems from API metadata, abstracts, or resolver records
        - only follow candidate URLs from APIs when the destination still matches the same DOI or the same exact title
     4. If the URL is a workshop/problem-session/forum page whose own body contains the relevant problem statements, use the page itself as the primary source.
        - examples: AIM pages, workshop problem-session pages, MathOverflow threads, Stack Exchange sites (e.g., Physics SE, Biology SE, Cross Validated), domain-specific Q&A forums
        - for Q&A forum sources, prefer the question body as the primary source; use answers/comments only for explicit resolution claims or clarifications that are needed and clearly attributable
     5. If direct fetching fails because of redirects, robots/SSL issues, transient HTML fetch failures, or citation-only landing pages, search for the same source by exact title, DOI, repository identifier, or trailing URL slug and recover a readable artifact for that same work/source. Do not switch to a different work.
   - If the resolved primary source is HTML rather than PDF, set:
     - `source_kind = "html"`
     - `source_locator = <canonical URL of the chosen page>`
     - `source_name = <arxiv_id_or_doi_slug_or_url_slug>`
     - `citation_mode = "locator_string"`

## Source Metadata

- Every final JSON object must include a top-level `source` object.
- `source.url` must preserve the original user-supplied source locator:
  - if input was a URL, store that original URL
  - if input was a local PDF path, store the absolute local path
- `source.resolved_locator` must record the canonical URL or local file path actually used for extraction.
- `source.title` must be a non-empty source title.
- `source.title_origin` must be one of:
  - `source_text`
  - `resolver_metadata`
  - `ai_extracted`
  - `filename`
- Determine `source.title` in this order:
  1. Prefer the visible title from the chosen primary source itself:
     - PDF title page, first page heading, or explicit article/workshop title
     - HTML `<title>`, `<h1>`, main heading, or obvious visible page title
  2. If the chosen primary source lacks a reliable visible title, use trusted resolver metadata for the same exact work/source.
  3. If no reliable title is available, infer a concise descriptive title from the readable source body and set `title_origin = "ai_extracted"`.
  4. If the source is unreadable and no better title is available, fall back to the local filename stem or URL slug and set `title_origin = "filename"`.
- Never leave `source.title` blank.

## Source Resolution Rules

- Prefer the direct PDF of the exact source document when available.
- Otherwise use full HTML text whose visible body actually contains the problem statements.
- Otherwise use the closest source-authorized readable copy.
- Treat direct-fetch failures such as robots-blocked pages, broken TLS, redirect loops, and citation-only HTML as resolver failures, not as proof that the source is unusable.
- Metadata APIs may be used for discovery, title recovery, DOI verification, and access-status checks, but never as the technical source for extraction.
- If a resolver or API confirms the record is metadata-only or restricted with no readable full text (for example a HAL `notice` record with `openAccess_bool=false`), stop retrying equivalent resolver URLs and move the item to `needs_review` unless you can recover another exact-work readable copy.
- Candidate links discovered in resolver metadata, API fields, or publisher abstracts must be treated as hints only; follow them only if the resulting page or file still matches the same DOI or the same exact title.
- Do not rely on abstract-only pages, citation-only pages, search snippets, or index metadata as the technical source when they do not contain the problem statements themselves.
- Do not combine multiple separate works into one run. If a resolver page exposes multiple editions, choose one canonical readable source and stay with it.

## Access-Failure Policy

For batch runs, resolve access problems with a bounded recovery pass. Prefer a clear `error_input` checkpoint over a long speculative search.

When a page has access issues, classify the failure mode first:

- `network_or_dns_failure`: shell/browser fetch cannot resolve or connect
- `blocked_or_challenged`: Cloudflare, JavaScript/cookie challenge, 403/401, robot block, or login wall
- `paywalled_or_restricted`: publisher landing page or resolver confirms no readable full text
- `metadata_only`: resolver, abstract, citation page, table of contents, index entry, or review page with no problem statements
- `missing_or_invalid_artifact`: download endpoint returns missing-resource HTML, a corrupt file, or a non-readable PDF
- `ambiguous_resolution`: multiple plausible works are found and none can be confirmed as the same exact source

Bounded recovery steps:

1. If the source has a stable identifier, try identifier-specific recovery first: arXiv PDF, DOI landing page, Crossref, OpenAlex, HAL API, zbMATH metadata, or explicit resolver links.
2. If metadata provides a DOI, exact title, year, and authors, use them only to find a readable artifact for that same work.
3. Run at most three exact-work search attempts after API/resolver recovery. Use precise queries such as exact DOI, exact title plus `PDF`, exact title plus first author, or repository identifier. Avoid broad topical searches.
4. Inspect at most five candidate pages or files from those searches. Accept a candidate only when the visible title or DOI confirms the same exact work/source.
5. Stop immediately once remaining candidates are only snippets, abstracts, citation metadata, paywalls, blocked pages, or different works.

Failure output rules:

- Do not extract questions from metadata, abstracts, search snippets, title text, or related papers.
- Do not keep retrying equivalent blocked resolver URLs once the same failure mode is established.
- If no readable exact-work PDF or full HTML source is recovered within the bounded pass, return the standard `error_input` JSON with `accepted: []`.
- Set `source.title` from source text when available, otherwise trusted resolver/API metadata, otherwise the URL slug or filename.
- Leave `source.resolved_locator` empty unless a readable primary source artifact was actually used for extraction.
- In `trace`, include the failure mode, identifiers attempted, and why the recovered candidates were rejected.

## Source-Family Hints

- **AIM / AIMPL pages**: if the exact URL does not fetch cleanly, search the AIM Problem Lists site by the trailing slug and exact visible title, then use the matching list page that contains the problem statements. If the root page is only an index, follow the numbered internal subpage that actually lists the problems.
- **Legacy HAL hosts**: normalize old `hal-*.ccsd.cnrs.fr` pages to their canonical `hal.science` record when possible; if needed, search by the HAL identifier and prefer a `document` or direct file URL. If `/document` returns a missing-resource page, query the HAL API by identifier. Treat records such as `submitType_s="notice"` with `openAccess_bool=false` as metadata-only dead ends unless another exact-work readable copy is found.
- **DBLP / zbMATH / OpenAlex / Crossref**: treat these as bibliographic resolvers. Prefer explicit links labeled `unpaywalled version`, `electronic edition`, `external edition`, `PDF`, `document`, `arXiv`, or DOI targets that lead to readable full text. If the resolver page still exposes only metadata, use the exact title, DOI, year, and author metadata to recover an accessible copy of that same work.
- **OpenAlex**: prefer `best_oa_location.pdf_url`, then `best_oa_location.landing_page_url`, then `primary_location` when they correspond to the same exact work. If `open_access.is_oa=false` and there is no repository or PDF location, treat OpenAlex as identity metadata only.
- **Crossref**: use Crossref to recover the exact title, DOI, and candidate resource URLs. Treat `resource.primary.URL`, `link[]`, and any full-text URL embedded in the abstract only as candidate leads, and only follow them when the resolved destination still matches the same DOI or exact title.
- **DOI URLs**: if the DOI landing page is paywalled or metadata-only, query Crossref and OpenAlex by DOI before doing a generic exact-title search. Prefer repository copies, accepted manuscripts, or author-posted PDFs of the same work.
- **zbMATH**: if the zbMATH page itself is access-denied or metadata-only, use the zbMATH identifier plus the exact title/year/author metadata to search for the same work elsewhere. If no readable exact-work copy exists, move the item to `needs_review` rather than extracting from bibliographic metadata.

If input is invalid or cannot be downloaded, return:

```json
{
  "source": {
    "url": "<original_input_url_or_path>",
    "resolved_locator": "",
    "title": "<best_effort_slug_or_filename>",
    "title_origin": "filename"
  },
  "accepted": [],
  "needs_review": [
    {
      "id": "error_input",
      "reason": "invalid_or_unreadable_input",
      "question_text_raw": "",
      "evidence": []
    }
  ],
  "trace": [
    {
      "id": "error_input",
      "stage": "input",
      "notes": "Input could not be normalized to a readable PDF or HTML full-text source.",
      "evidence_refs": []
    }
  ]
}
```

## Workflow Stages

### Stage 1: Map

- If `source_kind = "pdf"`, use `look_at(file_path=paper_path, goal=...)` to extract:

  - section hierarchy (major sections and subsections)
  - notation/terminology/definitions index
  - key result/claim/finding reference index (theorems, propositions, lemmas, experiments, algorithms, models, case studies, and similar formal structures, with labels and where they appear)
  - likely "open problems/questions" regions

- If `source_kind = "html"`, use the available web/page-reading tools on `source_locator` to extract:
  - section hierarchy or page structure (headings, numbered problem blocks, post/answer/comment blocks, list items)
  - notation/terminology/definitions index when present
  - likely "open problems/questions" regions
  - stable locators for later citation, such as section titles, numbered items, and HTML line spans

Record this as `doc_map` in working memory and add a `trace` entry with stage `map`.

### Stage 2: Candidate Extraction (Recall-First)

Iterate through all the sections in `doc_map` and extract ALL explicit and implicit open problems/questions.

Extraction target:

- include markers like "Question", "Problem", "Open", "Unknown whether", "Is it true that"
- include bullet lists, numbered problem sessions, named open-problem blocks, and forum-style top-level question statements when they are clearly research questions in the field
- include entries labeled as solved (e.g., `(Solved) problem ...`) and keep them in output
- include implicit question statements that represent unresolved research problems
- for forum-style sources, extract the main post and any explicitly separated subquestions; do not mine side comments or answers as separate research problems unless they clearly formulate one
- for each candidate, capture:
  - raw question text (as close to source as possible)
  - page number(s) or stable HTML locator(s)
  - verbatim supporting quote(s)
  - solved-status metadata derived from source labels (for example `(Solved)`, `Solved`, or equivalent)

Add one `trace` entry per section with stage `extract_candidates`.

### Stage 3: Self-Contained Rewrite (Precision-First)

For each candidate:

- rewrite into a standalone question understandable without the source
- `question_text` is the primary payload and must by itself be fully self-contained for a reader who has not seen the source or any other extracted question
- `question_text` should be as long as needed to inline all notation, terminology, definitions, assumptions, experimental conditions, datasets, models, evaluation metrics, domain context, and other problem data required for a fully usable standalone statement
- for extracted questions, prefer a clearly longer rewrite than the original source wording whenever that wording is too compressed to stand alone; the goal is faithful expansion, not brevity
- prefer over-explaining rather than under-explaining when deciding whether to inline definitions; if a careful reader could not start solving without source-specific terminology being unpacked, unpack it in `question_text`
- `question_text` may be a multi-sentence or multi-paragraph JSON string, but it must still read as one self-contained question rather than disconnected notes
- avoid source-pointing text in `question_text` such as "in the paper", "in this section", "as defined above", "the authors define", or similar phrasing that gestures at the source instead of stating the needed content directly
- `context_brief` is only a short label and must never carry definitions that are required to understand the problem
- resolve references like "Proposition 2.1", "Experiment 3", "Table 2", "Algorithm 1", "Figure 5", "Section 3", "Problem 4", or "the accepted answer" only using evidence in the chosen primary source artifact
- if required context is missing in evidence, do NOT guess; move item to `needs_review`

Add one `trace` entry per candidate with stage `rewrite`.

### Stage 4: Dedupe

- merge near-duplicate rewritten questions
- preserve all evidence references from merged members in `trace`
- keep one canonical `id` per merged cluster

Add one `trace` entry per candidate with stage `dedupe`.

### Stage 5: Quality Gates

Run deterministic gates:

1. Self-containedness gate:
   - no dangling references like "this", "above", "as discussed", "see Section X" unless expanded
   - if a reader would need source-specific notation, terminology, definitions, experimental conditions, datasets, evaluation criteria, or domain-specific context to understand the question, include them directly in `question_text` when supported by evidence
   - `question_text` must not rely on source-pointing language such as references to the paper, section, source text, or prior definitions instead of restating the content directly
2. Evidence coverage gate:
   - accepted items must include at least one evidence quote with a page number or stable HTML locator
3. Evidence completeness gate
   - accepted items must include complete sentences or a contiguous self-contained problem statement as evidence; otherwise, move them to `needs_review`
4. No-new-facts gate:
   - rewritten content, including any added definitions or settings in `question_text`, must not introduce unsupported claims
5. Source adequacy gate:
   - do not accept items extracted only from abstracts, citation metadata, link labels, thread titles, or search snippets when the technical problem statement is not visible in the chosen source text
6. Schema conformance gate:
   - accepted items must match the JSON output schema exactly and contain all required fields

Any failed item goes to `needs_review` with specific `reason`.
Add one `trace` entry per candidate with stage `quality_gates`.

## Required Prompt Clauses

## EVIDENCE_CITATION

- Every extracted or rewritten question must be backed by verbatim quote evidence and a source locator.
- Evidence format: `{ "page": <number_or_string>, "quote": "<verbatim text>" }`
- Use numeric page numbers for PDFs.
- Use stable locator strings for HTML/page/thread sources, such as `"HTML lines 57-58"`, `"Problem 3, HTML lines 23-28"`, or `"Accepted answer lines 167-169"`.
- If evidence cannot be located, do not accept the item.

## JSON_SERIALIZATION

- The final payload must be valid JSON that parses without repair.
- Escape every backslash inside JSON strings. This matters especially for sources with domain-specific markup such as TeX/LaTeX (mathematics), chemical formulas, code snippets, pseudocode, or structured notation (e.g., MathOverflow, arXiv HTML, bioRxiv).
- In `question_text`, prefer plain-text or Unicode notation over raw domain-specific markup (e.g., TeX/LaTeX, chemical notation, pseudocode) when either would be equally faithful.
- In `evidence.quote`, keep the source text verbatim, but serialize it as a valid JSON string with required escaping.
- Before returning and before saving, check that the exact payload would parse as JSON; if not, repair the escaping first.

## NO_NEW_FACTS

- Do not invent facts, assumptions, constraints, or definitions.
- Do not infer missing technical statements beyond provided evidence.
- When expanding a question to make it self-contained, only inline definitions, notation, assumptions, or ambient-setting details that are directly supported by the chosen source.
- Do not treat citation metadata, abstracts, or search snippets as sufficient technical evidence unless the chosen source page itself states the full problem there.
- If missing context is required to make the question self-contained, move to `needs_review`.

## NEEDS_REVIEW_FLAG

Move candidate to `needs_review` when any of these hold:

- unresolved cross-reference (proposition, theorem, experiment, figure, table, algorithm, or section not recoverable from evidence)
- ambiguous pronoun/deixis with unclear antecedent
- insufficient quote evidence for a rewritten claim
- undefined source-specific terminology remains in `question_text`
- `question_text` points back to the source instead of stating the needed content directly
- potential merge conflict between similar but distinct problems
- source resolves only to abstract/citation metadata without the full problem text
- resolver/API metadata confirms the record is restricted, notice-only, or lacks any readable full-text copy of the same work
- forum/workshop page is not self-contained enough and missing context cannot be recovered from the same source
- resolver chain is ambiguous between multiple possible full-text sources

## SOLVED_STATUS_METADATA

- Do not drop candidates just because they are labeled solved.
- Every accepted question row must include solved-status metadata.
- Metadata format in each accepted row:
  - `"meta": { "is_solved": <boolean> }`
- Set `is_solved=true` only when the chosen primary source explicitly marks the item as solved or explicitly resolves it in the same source artifact.
- Otherwise set `is_solved=false`.

## Output Schema (STRICT JSON)

Return exactly one JSON object:

```json
{
  "source": {
    "url": "https://example.org/problem-page",
    "resolved_locator": "https://example.org/problem-page",
    "title": "Example Problems Page",
    "title_origin": "source_text"
  },
  "accepted": [
    {
      "id": "q_001",
      "question_text": "Fully self-contained open question text. It may be substantially longer than the original source wording when needed to include notation, terminology, definitions, assumptions, experimental conditions, datasets, evaluation criteria, and domain context required to understand the problem without the source.",
      "context_brief": "Short topic label only; not a place to hide required definitions.",
      "meta": {
        "is_solved": false
      },
      "evidence": [
        {
          "page": 12,
          "quote": "Verbatim supporting quote from the source."
        }
      ]
    }
  ],
  "needs_review": [
    {
      "id": "q_017",
      "reason": "unresolved_cross_reference",
      "question_text_raw": "Original extracted statement with unresolved reference.",
      "evidence": [
        {
          "page": "HTML lines 57-58",
          "quote": "Verbatim source quote for the unresolved item."
        }
      ]
    }
  ],
  "trace": [
    {
      "id": "q_001",
      "stage": "rewrite",
      "notes": "Expanded Proposition 2.1 reference using evidence from page 10.",
      "evidence_refs": [
        "p10",
        "p12"
      ]
    }
  ]
}
```

Do not include markdown fences in final answer. Output raw JSON only.

`page` may be either a numeric page number or a stable string locator, depending on `citation_mode`.

## Auto-Save (MANDATORY)

Before returning the final JSON to user:

1. Create output directory:
   - `RUN_DIR`
2. Generate output file name from the normalized source name:
   - `<source_name_sanitized>.json` (local PDF stem, arXiv id, DOI slug, or URL slug)
   - Sanitization rule: replace every `/` with `_` and every remaining non-alphanumeric character except `-` and `.` with `_`; do not use `-` as a slash replacement
3. Save the exact JSON output to:
   - `RUN_DIR/<source_name_sanitized>.json`
4. Also write/update:
   - `LATEST_FILE`
5. Append a `trace` entry noting save paths under stage `persist`.

If save fails, still return JSON but add a `needs_review` item with reason `persist_failed` and include failure details in `trace`.
\end{tcblisting}
\begin{center}
\captionsetup{type=table,hypcap=false}
\caption{Prompt for the Extractor Agent}
\label{tab:extractor_prompt}
\end{center}

\begin{tcblisting}{
  enhanced,
  breakable,
  listing only,
  colback=teal!04,
  colframe=teal!25,
  coltitle=black,
  fonttitle=\small\bfseries\color{black},
  sharp corners,
  boxrule=0.6pt,
  left=1.2mm,right=1.2mm,top=1.0mm,bottom=1.0mm,
  before skip=6pt, after skip=2pt,
  title={System and User Prompts for the Refiner Agent},
  listing options={
    basicstyle=\ttfamily\scriptsize,
    breaklines=true,
    breakatwhitespace=false,
    columns=fullflexible,
    keepspaces=true,
    showstringspaces=false,
    escapeinside={(*@}{@*)}
  }
}
(*@\textbf{System prompt}@*)
You are a science research assistant that refines draft research questions by consulting the original arXiv paper.

(*@\textbf{User prompt}@*)
I will give you an arXiv paper ID and a draft question.

Your job is to read the paper itself and turn my draft into a highly complete, sharp, stand-alone research question that stays as close as possible to the paper's actual intent.

Paper to inspect:
arXiv:{arxiv_id}

Draft question:
{question_text}

Instructions:

1. Search the paper and identify the exact statement, conjecture, or open problem that my draft is trying to capture.
   - Prefer the nearest official formulation in the paper.
   - Do not replace it with a broader neighboring problem.

2. Read the exact statement and the nearby context that defines notation, scope, and hidden assumptions.
   - Check the introduction first.
   - Then check any proposition/theorem/conjecture/remark/hypothesis that supplies equivalent conditions, notation, or a more precise reformulation.

3. Evaluate how complete my draft question is on a 0 to 10 scale.
   - Say what is already correct.
   - List every missing or implicit detail needed to make it fully stand-alone and paper-faithful.
   - Include domain-specific definitions (e.g. variables, operators, experimental conditions, model assumptions, scope restrictions, quantifiers) that a reader outside the paper would need to understand and attempt the question.

4. Rewrite the question as one refined research question that is:
   - self-contained (no unexplained symbols, jargon, or implicit scope),
   - tightly scoped,
   - as close as possible to the paper's original intent,
   - not broader than the paper,
   - not vague.

5. Give at most one alternate formulation only if the paper itself explicitly proves the equivalence.
   - Do not invent new variants.
   - Do not generalize unless the paper itself does so in the exact nearby discussion.

6. Guardrails:
   - Do not broaden a specific narrow question into a more general problem unless the paper explicitly treats that generalization as its main target.
   - Do not drift into side problems, variants, or adjacent subfields unless my draft explicitly asks for them.
   - Do not add speculative assumptions or remove paper-critical restrictions (e.g. specific parameter ranges, experimental constraints, model assumptions).
   - Keep the result narrow.
   - The "refined_question" field must contain only plain text. Do not include any URLs, hyperlinks, DOIs, PDF references, arXiv links, or citations of any kind inside the refined question itself.

7. Support every nontrivial claim by citing the exact place in the paper:
   - section number,
   - conjecture/proposition/theorem/open problem number,
   - and page number if available.

Output requirements:

Return ONLY a valid JSON object.

The JSON must contain the following fields:

- "target_in_paper": string
- "completeness_score": number
- "correct_elements": string
- "missing_details": array of strings
- "refined_question": string
- "equivalent_formulation": string or null
- "justification": string
- "citations": array of objects with:
    - "section": string
    - "label": string
    - "page": string or number

Do NOT include any text outside the JSON.
Do NOT include markdown formatting.
Ensure the JSON is syntactically valid.
\end{tcblisting}
\begin{center}
\captionsetup{type=table,hypcap=false}
\caption{System and User Prompts for the Refiner Agent}
\label{tab:refiner_prompt}
\end{center}

\begin{tcblisting}{
  enhanced,
  breakable,
  listing only,
  colback=teal!04,
  colframe=teal!25,
  coltitle=black,
  fonttitle=\small\bfseries\color{black},
  sharp corners,
  boxrule=0.6pt,
  left=1.2mm,right=1.2mm,top=1.0mm,bottom=1.0mm,
  before skip=6pt, after skip=2pt,
  title={Prompt for Difficulty Comparison},
  listing options={
    basicstyle=\ttfamily\scriptsize,
    breaklines=true,
    breakatwhitespace=false,
    columns=fullflexible,
    keepspaces=true,
    showstringspaces=false
  }
}
You are an expert mathematics judge comparing two math problems.

Your task is to decide which problem is more difficult in each dimension below.
A "win" means the problem is more difficult for that dimension; a "lose" means it is less difficult.
Use the labels "problem_1" and "problem_2" exactly. Use "draw" when neither problem is meaningfully more difficult in that dimension.
Do not solve the problems fully. Reason only enough to compare relative difficulty.

Return only a valid JSON object with exactly these top-level keys:
- procedural_difficulty
- memory_or_knowledge_difficulty
- novelty_difficulty

Each value must be an object with this exact shape:
{{
  "winner": "problem_1, problem_2, or draw",
  "problem_1": "win, lose, or draw",
  "problem_2": "win, lose, or draw",
  "reason": "one concise sentence"
}}

Example output format:
{{
  "procedural_difficulty": {{
    "winner": "problem_2",
    "problem_1": "lose",
    "problem_2": "win",
    "reason": "Problem 2 requires more algebraic manipulation and case bookkeeping."
  }},
  "memory_or_knowledge_difficulty": {{
    "winner": "draw",
    "problem_1": "draw",
    "problem_2": "draw",
    "reason": "Both problems require a comparable level of standard background knowledge."
  }},
  "novelty_difficulty": {{
    "winner": "problem_2",
    "problem_1": "lose",
    "problem_2": "win",
    "reason": "Problem 2 hides the relevant method behind a less familiar formulation."
  }}
}}

Problem 1:
{problem_1}

Problem 2:
{problem_2}

Dimensions:

## Procedural difficulty

How many steps or calculations are required.

A problem can be conceptually easy but procedurally long.

Example: expanding a large polynomial, doing long algebra, or computing a determinant by hand.

This includes:

| Aspect | Meaning |
|---|---|
| Number of steps | More steps means more chances to make errors |
| Algebraic manipulation | Simplifying, rearranging, expanding, factoring |
| Calculation burden | Arithmetic, fractions, signs, exponents |
| Bookkeeping | Keeping track of cases, variables, or expressions |

## Memory or knowledge difficulty

How much prior knowledge the solver needs.

Example: a problem may be easy if you know the AM-GM inequality, but very hard if you do not.

This includes:

| Aspect | Meaning |
|---|---|
| Formula recall | Remembering identities or standard formulas |
| Theorem recall | Knowing which results are available |
| Technique familiarity | Having seen similar patterns before |
| Domain-specific knowledge | Number theory, combinatorics, topology, etc. |

## Novelty difficulty

How different the problem is from familiar examples.

A routine integral may be easy. A strange-looking integral requiring an unusual substitution may be hard.

This includes:

| Aspect | Meaning |
|---|---|
| Familiarity | Whether the solver has seen similar problems |
| Surface disguise | Familiar idea hidden under unusual wording |
| Creative leap | Need to invent or adapt a method |
| Transfer | Applying a known method in a new context |
\end{tcblisting}
\begin{center}
\captionsetup{type=table,hypcap=false}
\caption{Prompt for Difficulty Comparison}
\label{tab:difficulty-comparision-prompt}
\end{center}

\begin{tcblisting}{
  enhanced,
  breakable,
  listing only,
  colback=teal!04,
  colframe=teal!25,
  coltitle=black,
  fonttitle=\small\bfseries\color{black},
  sharp corners,
  boxrule=0.6pt,
  left=1.2mm,right=1.2mm,top=1.0mm,bottom=1.0mm,
  before skip=6pt, after skip=2pt,
  title={Prompt for Response Generation},
  listing options={
    basicstyle=\ttfamily\scriptsize,
    breaklines=true,
    breakatwhitespace=false,
    columns=fullflexible,
    keepspaces=true,
    showstringspaces=false
  }
}
Solve the following problem.

Problem:
{problem}
\end{tcblisting}
\begin{center}
\captionsetup{type=table,hypcap=false}
\caption{Prompt for Response Generation}
\label{tab:response-prompt}
\end{center}

\begin{tcblisting}{
  enhanced,
  breakable,
  listing only,
  colback=teal!04,
  colframe=teal!25,
  coltitle=black,
  fonttitle=\small\bfseries\color{black},
  sharp corners,
  boxrule=0.6pt,
  left=1.2mm,right=1.2mm,top=1.0mm,bottom=1.0mm,
  before skip=6pt, after skip=2pt,
  title={Committed-Attempt System Prompt},
  listing options={
    basicstyle=\ttfamily\scriptsize,
    breaklines=true,
    breakatwhitespace=false,
    columns=fullflexible,
    keepspaces=true,
    showstringspaces=false
  }
}
You are a bold research mathematician. Your task is to produce a COMPLETE, COMMITTED attempt at solving the problem, a full proof, construction, or explicit answer ending in a definite conclusion, even when you are not sure it is correct. A confident, fully-worked attempt that turns out to be WRONG is exactly what is wanted here; it is far more valuable than a cautious survey, a hedge, or an admission that the problem is unsolved. Do not optimize for being right; optimize for a complete, concrete line of reasoning that reaches a definite answer.

HARD RULES:
- COMMIT to a definite answer. Yes/no question -> pick one and prove it. Classification/construction/value -> give the explicit one. If you don't know, GUESS, then argue for the guess in full as if true.
- NEVER conclude the answer is 'open', 'unknown', 'not known', 'still unresolved', or 'beyond current methods'. Those endings are forbidden. Uncertainty is a reason to commit to your best guess, not to hedge.
- Do NOT survey the literature in place of solving. Prove something yourself.
- SOLVE THE EXACT PROBLEM, NOT AN EASIER ONE. First restate the exact problem in one line, listing every hypothesis, quantifier, and constraint. Solve that. Do not drop, weaken, or add a hypothesis, change 'for all' to 'there exists', or replace the general case with a special case. If you catch yourself proving something easier, stop and return to the stated problem.
- NAME THE CRUX, DON'T DISGUISE IT. Identify the single hardest step, labeled 'CRUX:'. Either prove it in full, or commit to it as an explicit 'ASSUMPTION (unproven):' and build on it. Guessing the crux is fine; presenting an unproven bridging step as an established lemma, a 'standard result', 'well known', or a finished proof is forbidden.
- A reduction or 'generalization' counts only if it is genuinely as hard or as broad as the original; if it isn't, say so rather than claiming progress you didn't make.
- Write the complete final solution with a clearly-marked final answer, and end with one line: "Proved: ... / Assumed without proof: ...".

You may be wrong. That is fine and expected. But solve the problem you were actually given, and be honest about which steps are proved versus assumed. Produce the strongest, most complete attempt you can and COMMIT to it.
\end{tcblisting}
\begin{center}
\captionsetup{type=table,hypcap=false}
\caption{Committed-Attempt System Prompt for Response Generation}
\label{tab:committed-attempt-prompt}
\end{center}

\begin{tcblisting}{
  enhanced,
  breakable,
  listing only,
  colback=teal!04,
  colframe=teal!25,
  coltitle=black,
  fonttitle=\small\bfseries\color{black},
  sharp corners,
  boxrule=0.6pt,
  left=1.2mm,right=1.2mm,top=1.0mm,bottom=1.0mm,
  before skip=6pt, after skip=2pt,
  title={Prompt for Factuality Metric: Reference Span Extraction},
  listing options={
    basicstyle=\ttfamily\scriptsize,
    breaklines=true,
    breakatwhitespace=false,
    columns=fullflexible,
    keepspaces=true,
    showstringspaces=false
  }
}
You are an exact-span annotation tool. Return only a JSON array of strings.

Task: identify external references in one block from a mathematical reasoning trace.

Definition:
An external reference is any arXiv/paper/book mention, website/URL, named author/source used as support.

Examples of external references:
- "the paper by Smith et al."
- "according to Wikipedia"
- "from the Art of Problem Solving wiki"
- "in the book 'Affine Algebraic Geometry' by Gutierrez, Masuda, etc."

DO NOT extract:
- Theorems or lemmas with names, such as the Cauchy-Schwarz inequality, Descartes' theorem, or Burnside's lemma
- mathematical objects, variables, equations, or theorems used without an outside source
- references to prior reasoning attempts

Rules:
1. Copy the complete exact substring from the block for each external reference.
2. Every string must be a contiguous substring that appears verbatim in the block.
3. Do not paraphrase, normalize spelling, add positions, or add text that is not in the block.
4. Do not output overlapping strings. If one citation contains another marker, output only the larger complete citation/source phrase.
5. If there is no external reference, output [].
6. Output JSON only. No Markdown, no explanation.


Block:
<<<BLOCK
{block}
BLOCK
\end{tcblisting}
\begin{center}
\captionsetup{type=table,hypcap=false}
\caption{Prompt for factuality reference-span extraction}
\label{tab:fact-first-stage-prompt}
\end{center}

\begin{tcblisting}{
  enhanced,
  breakable,
  listing only,
  colback=teal!04,
  colframe=teal!25,
  coltitle=black,
  fonttitle=\small\bfseries\color{black},
  sharp corners,
  boxrule=0.6pt,
  left=1.2mm,right=1.2mm,top=1.0mm,bottom=1.0mm,
  before skip=6pt, after skip=2pt,
  title={Prompt for Factuality Metric: Agent Verification},
  listing options={
    basicstyle=\ttfamily\scriptsize,
    breaklines=true,
    breakatwhitespace=false,
    columns=fullflexible,
    keepspaces=true,
    showstringspaces=false
  }
}
You are verifying one detected external-reference span from a mathematical reasoning trace.

Important:
- You MUST use web search before deciding.
- Return exactly one lowercase boolean token: true or false.

Decision rule:
- Return true only if the target is a purported specific external reference or citation
  (for example a named paper, book, author/person used as a source, website/page, or
  similar source) and web search indicates that this cited reference is likely
  fabricated, nonexistent, or materially mismatched.
- Return false if the target is a real existing reference/source/author/website/page.
- Return false if the target is not actually an external reference, such as a theorem,
  concept, generic phrase, internal note, or ordinary problem text.
- If the evidence is ambiguous and the target is not clearly a fabricated specific
  citation, return false.

Remember: true means "keep this as a fake reference"; false means "filter it out".

Problem:
<<<PROBLEM
{problem_text}
PROBLEM

Target Reference:
<<<REFERENCE
{reference}
REFERENCE

Exact Detected Block:
<<<EXACT_BLOCK
{block_text}
EXACT_BLOCK
\end{tcblisting}
\begingroup
\captionsetup{type=table,hypcap=false,position=bottom,skip=8pt}
\caption{Prompt for factuality agent verification}
\label{tab:fact-second-stage-prompt}
\endgroup

\par
\noindent\begin{minipage}{\linewidth}
\begin{tcblisting}{
  enhanced,
  breakable,
  listing only,
  colback=teal!04,
  colframe=teal!25,
  coltitle=black,
  fonttitle=\small\bfseries\color{black},
  sharp corners,
  boxrule=0.6pt,
  left=1.2mm,right=1.2mm,top=1.0mm,bottom=1.0mm,
  before skip=6pt, after skip=2pt,
  title={Short-Block Prompt for Factuality Metric: Agent Verification},
  listing options={
    basicstyle=\ttfamily\scriptsize,
    breaklines=true,
    breakatwhitespace=false,
    columns=fullflexible,
    keepspaces=true,
    showstringspaces=false
  }
}
You are verifying one detected external-reference span from a mathematical reasoning trace.

Important:
- You MUST use web search before deciding.
- Return exactly one lowercase boolean token: true or false.

Decision rule:
- Return true only if the target is a purported specific external reference or citation
  (for example a named paper, book, author/person used as a source, website/page, or
  similar source) and web search indicates that this cited reference is likely
  fabricated, nonexistent, or materially mismatched.
- Return false if the target is a real existing reference/source/author/website/page.
- Return false if the target is not actually an external reference, such as a theorem,
  concept, generic phrase, internal note, or ordinary problem text.
- If the evidence is ambiguous and the target is not clearly a fabricated specific
  citation, return false.

Remember: true means "keep this as a fake reference"; false means "filter it out".

Problem:
<<<PROBLEM
{problem_text}
PROBLEM

Target Reference:
<<<REFERENCE
{reference}
REFERENCE

Exact Detected Block:
<<<EXACT_BLOCK
{block_text}
EXACT_BLOCK

Surrounding Reasoning Window Containing The Target:
<<<BLOCK
{prompt_context}
BLOCK
\end{tcblisting}
\begingroup
\captionsetup{type=table,hypcap=false,position=bottom,skip=3pt}
\caption{Short-block prompt for factuality agent verification}
\label{tab:fact-second-stage-short-block-prompt}
\endgroup
\end{minipage}

\end{document}